\pdfoutput=1

\documentclass[11pt]{article}

\usepackage[final]{acl}

\usepackage{times}
\usepackage{latexsym}

\usepackage[T1]{fontenc}

\usepackage[utf8]{inputenc}

\usepackage{times}
\usepackage{latexsym}
\usepackage{booktabs}
\usepackage{amsmath}
\usepackage{float}
\usepackage{adjustbox}
\usepackage{longtable}
\usepackage{xcolor,colortbl}
\usepackage{diagbox} 
\usepackage{array,multirow}
\usepackage{graphicx} 
\usepackage{placeins} 
\usepackage{microtype} 
\usepackage{enumitem} 
\usepackage{amsmath, bm} 
\usepackage{comment} 
\usepackage{amsmath, bm} 
\usepackage{subcaption}
\usepackage{transparent}
\usepackage{geometry}
\usepackage{pdflscape}
\definecolor{grey}{named}{gray}

\usepackage{microtype}

\usepackage{inconsolata}

\newcommand{\chl}[3]{\cellcolor{#1!#2}{#3}}  

\title{Preserving Multilingual Quality While Tuning Query Encoder on English Only}

\author{Oleg Vasilyev, Randy Sawaya, John Bohannon \\
  Primer Technologies Inc. \\
  San Francisco, California \\
  \texttt{{oleg,randy.sawaya,john}@primer.ai}\\}

\begin{document}
\maketitle
\begin{abstract}
A query encoder of a dual passage retrieval system can be tuned for specific types of queries or domains, while the precomputed and stored documents representations are kept intact. Switching from one query encoder to another when needed is easily feasible, unlike overhauling the embeddings of a whole knowledge base. In this work we raise a question: Can the generic, original qualities of the encoder be preserved or at least left not too degraded when it is tuned on a narrow domain? We conducted experiments on a high quality multilingual embedding model: Tuning it on a single English-only dataset, we observe that the tuning not only preserves the multilingual qualities, but even improves them. The embedding qualities on distinctly different data are also improved or at least preserved.
Drawing on our observations, we suggest a more general hypothesis: Tuning with intentionally low learning rate can preserve or improve a system's properties acquired in training, but not specifically targeted by tuning. We call this \textit{adiabatic tuning} and provide tentative explanations.
\end{abstract}

\section{Introduction}\label{sec:Introduction}
Advances in neural NLP methods have resulted in high quality dense vector text representations \cite{reimers-gurevych-2019-sentence, cer2018universal, conneau-etal-2017-supervised}. Such representations are often used at the initial stages of an information retrieval system, selecting the most relevant documents, ranked relative to the query \cite{xiong2020approximate, zhan2020repbert, 10.1145/3404835.3462880, ren-etal-2021-rocketqav2}. A dual encoder is successfully used to train the representations \cite{karpukhin-etal-2020-dense, ren-etal-2021-pair, qu-etal-2021-rocketqa, 10.1145/3404835.3462891, ni-etal-2022-large, dong-etal-2022-exploring}. A dual encoder dense passage retrieval system is efficient for two main reasons: (1) it allows using the simple inner product of query and document representations, and (2) it allows modifying the query representation for a task or domain, while keeping the stored and precomputed (query-invariant) document representations intact.

If the representation was pretrained in a multilingual setting, tuning on English-only samples may be expected to degrade the multilingual qualities and there may not be enough cross-lingual samples for tuning on a specific domain or types of queries. 
A multilingual query generator may be employed to overcome a shortage of cross-lingual data \cite{ren-etal-2022-empowering, 10.1145/3539618.3591952}, but, in this work, we follow an arguably simpler strategy. In order to understand the effect of English-only tuning on multilingual qualities of a representation, and to assess a possible degradation, we consider a simple setup: A state of the art multilingual embedding model is taken as the starting point, and fine-tuned by English only samples as the query representation part of a dual encoder. 

We assume that our observations of the degradation or preservation of the multilingual qualities may be generalized to other pretrained system qualities that are not directly targeted in tuning. In order to obtain preliminary confirmation of this hypothesis, we also observe the effect of tuning on the embedding quality for queries and text chunks of very different styles, the likes of which could be present in the training of the original encoder, but certainly not targeted in tuning.

Our contribution:
\begin{enumerate}[topsep=0pt,itemsep=-1ex,partopsep=1ex,parsep=1ex]
    \item We show that fine-tuning a query encoder on an English-only dataset may not only preserve multilingual qualities of query-document embeddings matching, but even improve them. 
    \item We hypothesize that a tuning regime with intentionally low learning rate (far below of what is necessary to avoid overfitting) preserves or improves the properties acquired in the training, but not targeted by tuning. We call this \textit{adiabatic tuning} and suggest supporting observations and conjectural explanations.
    \item We add a dataset with graded difficulty, based on ARXIV titles and abstracts.
\end{enumerate}
Although high-resource languages can be used for cross-lingual transfer~\cite{lin-etal-2019-choosing}, our setting does not have such a goal: the tuning is set to improve the query part of a dual encoder on a certain dataset, with no driving mechanism for preserving or improving the other qualities of the system.

Our starting point is one of the best (for its lean size) multilingual \underline{embedding} models which differs from starting with a multilingual \underline{language} model and then aligning the generated embeddings for different languages~\cite{wang2022englishcontrastivelearninglearn}.

\section{Setup}\label{sec:Setup}

\subsection{Models}\label{ssec:Model}
In what follows, we use a state-of-the-art multilingual model \textit{intfloat/multilingual-e5-small}\footnote{https://huggingface.co/intfloat/multilingual-e5-small} \cite{wang2024multilingual} which will be referred to here as $E5$. 
For most of the evaluations, we also consider results using \textit{sentence-transformers/paraphrase-multilingual-MiniLM-L12-v2}\footnote{https://huggingface.co/sentence-transformers/paraphrase-multilingual-MiniLM-L12-v2} \cite{reimers-gurevych-2019-sentence}, referred to as $L12$. Finally, we confirm some observations with monolingual \textit{intfloat/e5-small-v2}\footnote{https://huggingface.co/intfloat/e5-small-v2} \cite{wang2024textembeddingsweaklysupervisedcontrastive}, referred to as $E5e$.
All these models provide embeddings of a practical small size of $384$.

\subsection{Datasets}\label{ssec:Datasets}
We use MSMARCO \cite{nguen2018msmarco} Triplets\footnote{https://huggingface.co/datasets/sentence-transformers/embedding-training-data/blob/main/msmarco-triplets.jsonl.gz} for tuning and evaluation. For evaluating the qualities not targeted by tuning, we use the ARXIV dataset with negatives\footnote{https://huggingface.co/datasets/primer-ai/arxiv-negatives}, which we made from arxiv (version 173)\footnote{https://huggingface.co/datasets/arxiv-community/arxiv\_dataset}\footnote{https://www.kaggle.com/datasets/Cornell-University/arxiv}, and the test subset of the XNLI multilingual dataset\footnote{https://huggingface.co/datasets/facebook/xnli} \cite{conneau-etal-2018-xnli}. We also use HOTPOTQA\footnote{https://hotpotqa.github.io/} \cite{yang-etal-2018-hotpotqa} and SQUAD\footnote{https://huggingface.co/datasets/rajpurkar/squad\_v2} \cite{rajpurkar-etal-2018-know, rajpurkar-etal-2016-squad} for confirming some observations (Appendices~\ref{app:dataset_squad},~\ref{app:dataset_hotpotqa}).

Our test subset of MSMARCO contains 357642 evaluation triplets, made of 7000 samples - all the positives and negatives are used (Appendix~\ref{app:msmarco}).

Of ARXIV we use titles and abstracts. We made two flavors of evaluation arxiv triplets: (1) \textit{arxiv-title} where a title plays role of the query (anchor), and the corresponding abstract is a positive passage, and (2) \textit{arxiv-first} where the first sentence of abstract is used as the query, and the rest of it is used as a positive (Appendix~\ref{app:arxiv}).
We also use narrow versions of \textit{arxiv-first} in Appendix~\ref{app:narrow_tuning}.

\subsection{Tuning and evaluations}\label{ssec:Tuning}
Unless otherwise specified, we freeze the text encoder and proceed to fine-tune only the query encoder (fully or partially unfrozen) by contrastive learning on MSMARCO (or on narrow ARXIV subsets, Appendix~\ref{app:narrow_tuning}) with a learning rate of \mbox{5e-8}, batch size of 14 and the triple margin loss with margin $0.1$. Other details are in Appendix~\ref{app:tuning}.
In our experiments we considered different settings of freezing, batch size, learning rate, the margin of triplet loss, the stopping criterion, weight decay, scheduling versions and optimizers.

In most of our evaluations, we compare the similarity (or distance) between the anchor (query) and the positive vs the negative. If the positive does not turn out to be closer than the negative to the anchor, we count this as an error. We thus characterize performance of the encoder on a query by the number of errors divided by the total number of positive-negative pairs. We call this \textit{positive-negative discrepancy} (PND). The measure is easy to interpret, and its range (from 0 to 1) is the same and equally fair for any amounts of positives and negatives, as long as they exist in a selection for a query. On multiple queries we take an averaged PND. We confirm some results also using mean reciprocal rank (MRR), mean average precision (MAP) and precision at top 1 (P@1).
The improvement of performance is measured as relative change of a measure $M$ (PND or MRR or other):
\begin{equation} \label{eq:improvement}
I = s \frac{\tilde{M} - M}{M}
\end{equation}
where $M$ is for the original encoder, and $\tilde{M}$ is for the encoder after the tuning. The sign $s=-1$ for PND, because it decreases when improved, and $s=1$ for the other measures.   

For evaluating XNLI we use its pairs of sentences, each sentence is given in 15 languages (Appendix~\ref{app:xnli}). One sentence is used as a query, another as a passage. All pairs are human-labeled as entailment, neutral or contradiction. Hence, the sentences of an entailment pair should be closer to each other than the sentences of any neutral or contradiction pair. Whenever this does not happen, we count this as an error for PND.
In Appendix~\ref{app:performance_untuned} we made sure that the amount of errors the original encoder makes on our datasets is large enough to consider how tuning would affect them.

\section{Observations}\label{sec:Observations}

\begin{table*}[th!]
\centering
\begin{tabular}{@{}l|rr|rr|rr|ll|ll@{}}
\toprule
{} & \multicolumn{2}{c|}{msmarco} & \multicolumn{2}{c|}{arxiv-first} & \multicolumn{2}{c|}{arxiv-title} & \multicolumn{2}{c|}{xnli ent-neutr} & \multicolumn{2}{c}{xnli ent-contr}\\
{frozen}&{c\%}&{d\%}&{c\%}&{d\%}&{c\%}&{d\%}&{c+/-}&{d+/-}&{c+/-}&{d+/-}\\
\hline
    {-}&{7.47}&{8.46}&{5.19}&{5.19}&{\cellcolor{grey!10}{1.75}}&{\cellcolor{grey!10}{5.52}}&{222/0}&{215/2}&{194/4}&{147/23}\\
    
    {emb.base}&{7.32}&{8.82}&{4.85}&{5.41}&{\cellcolor{grey!10}{3.51}}&{\cellcolor{grey!10}{7.73}}&{222/0}&{217/1}&{201/2}&{159/21}\\
    
    {emb}&{7.30}&{8.82}&{4.90}&{5.39}&{\cellcolor{grey!10}{3.51}}&{\cellcolor{grey!10}{7.73}}&{222/0}&{217/1}&{201/2}&{158/20}\\
    
    {emb, B0a}&{7.30}&{8.76}&{4.77}&{5.34}&{\cellcolor{grey!10}{3.26}}&{\cellcolor{grey!10}{7.73}}&{222/0}&{217/1}&{200/2}&{159/21}\\
    
    {emb, B0a,i}&{7.48}&{9.00}&{5.05}&{5.36}&{\cellcolor{grey!10}{3.26}}&{\cellcolor{grey!10}{7.73}}&{223/0}&{219/0}&{199/2}&{156/25}\\
    
    {emb, B0a,i,od}&{7.31}&{8.82}&{4.77}&{5.19}&{\cellcolor{grey!10}{3.51}}&{\cellcolor{grey!10}{7.73}}&{222/0}&{217/1}&{200/2}&{158/21}\\
    
    {emb, B0}&{7.35}&{8.78}&{4.77}&{5.44}&{\cellcolor{grey!10}{3.51}}&{\cellcolor{grey!10}{7.73}}&{222/0}&{217/1}&{200/2}&{159/19}\\
    
    {emb, B0-5}&{7.87}&{9.39}&{5.79}&{6.07}&{\cellcolor{grey!10}{3.26}}&{\cellcolor{grey!10}{7.51}}&{219/0}&{213/3}&{200/5}&{157/25}\\
    
    {emb, B0-10}&{\cellcolor{grey!10}{1.45}}&{2.57}&{\cellcolor{grey!10}{0.89}}&{\cellcolor{grey!10}{1.21}}&{\cellcolor{grey!10}{0.00}}&{\cellcolor{grey!10}{0.44}}&{123/0}&{112/0}&{21/0}&{25/10}\\
\hline
\end{tabular}
\caption{Evaluations of the $E5$ query model tuned on MSMARCO as described in Section~\ref{ssec:Tuning}. The rows are in the order of increased freezing (at tuning): from no freezing (top row) to freezing everything up to the last transformer block $B11$.
The $emb.base$ model has only the first three layers of the embedding block frozen (tokens, positions, token-types). The $emb$ model has the full embedding block frozen. For the other notation: $B0$ is the full first transformer block; $B0$-$5$ are the first 6 blocks; the extensions $a$, $i$, $od$ (for $B0$) denote the layers
$attention$, $intermediate$ and $output.dense$ of the block. 
The columns $c\%$ and $d\%$ show the PND improvement (in percents) relative to the original model, accessed by cosine ($c$) or distance ($d$), grayed if not significant (Appendix~\ref{app:significance}). The columns $c$+/- and $d$+/- show count of language pairs with PND significantly improved ($+$) or worsened ($-$).}
\label{tab:improve_E5_freeze}
\end{table*}

\subsection{Tuning partially frozen query model}\label{ssec:freezing}
In Table~\ref{tab:improve_E5_freeze} we show results of tuning the dual encoder, with the text encoder frozen and query model free or partially frozen. 
Here and throughout the paper we use the easiest version of ARXIV (see Appendix~\ref{app:arxiv_difficulty_results} on performance at other levels).
Freezing the embedding block appears to be the best option for preserving the multilingual qualities, and henceforth it is used unless specified otherwise. In Table~\ref{tab:improve_E5default_ondatasets} we confirm the improvement on six other datasets (Appendices~\ref{app:msmarco},~\ref{app:dataset_squad},~\ref{app:dataset_hotpotqa}), and show some other measures.

\begin{table*}[th!]
\centering
\begin{tabular}{@{}l|rr|rr|rr|rr@{}}
\toprule
{} & \multicolumn{2}{c|}{PND} & \multicolumn{2}{c|}{MRR} & \multicolumn{2}{c|}{MAP} & \multicolumn{2}{c}{P@1}\\
{Dataset}&{c\%}&{d\%}&{c\%}&{d\%}&{c\%}&{d\%}&{c\%}&{d\%}\\
\hline
    {MSMARCO 65 negatives}&{2.41}&{3.78}&{0.48}&{0.55}&{1.03}&{1.15}&{1.92}&{2.02}\\
    {SQUAD}&{1.02}&{1.12}&{0.17}&{0.2}&{0.17}&{0.19}&{0.31}&{0.33}\\
    {SQUAD min 5}&{0.85}&{1.13}&{0.16}&{0.24}&{0.18}&{0.26}&{0.32}&{0.44}\\
    {HotpotQA easy}&{2.52}&{3.47}&{0.25}&{0.34}&{0.09}&{0.08}&{0.16}&{0.12}\\
    {HotpotQA medium}&{2.53}&{3.57}&{0.33}&{0.49}&{0.07}&{0.09}&{0.11}&{0.13}\\
    {HotpotQA hard}&{2.43}&{3.70}&{0.30}&{0.50}&{0.07}&{0.11}&{0.12}&{0.15}\\
\hline
\end{tabular}
\caption{Improvements for $E5$ tuned with frozen embedding block and learning rate 5e-8.}
\label{tab:improve_E5default_ondatasets}
\end{table*}

The multilingual qualities are not only preserved, but even mostly improved, especially on cosine similarity. The PND improvement is shown for each language pair separately in Figure~\ref{fig:hmap_xnli_gain_e5_c}. The results for the $L12$ model are similar (Appendix~\ref{app:L12_freezing}). In Appendix~\ref{app:narrow_tuning} we also also confirm our observations with $E5$ tuned on specific categories of ARXIV.

\begin{figure}[h!]
    \centering
    \includegraphics[width=\linewidth]{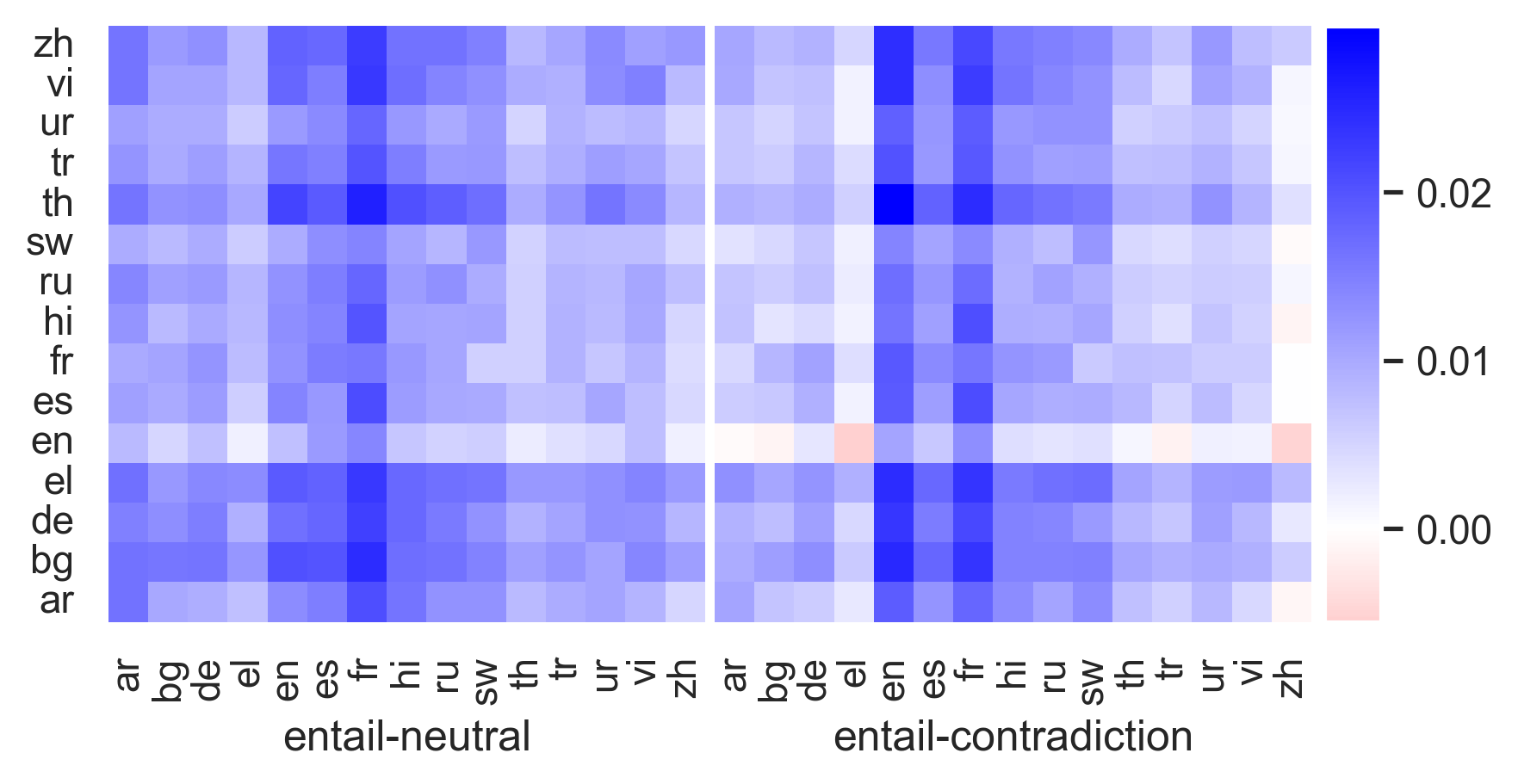}
    \caption{Improvement of $E5$ on XNLI assessed by cosine. Query is on axis $Y$; text is on $X$.}
    \label{fig:hmap_xnli_gain_e5_c}
\end{figure}

\subsection{Learning rate and adiabatic tuning}\label{ssec:adiabatic tuning}

\begin{figure}[h!]
    \centering
    \includegraphics[width=\linewidth]{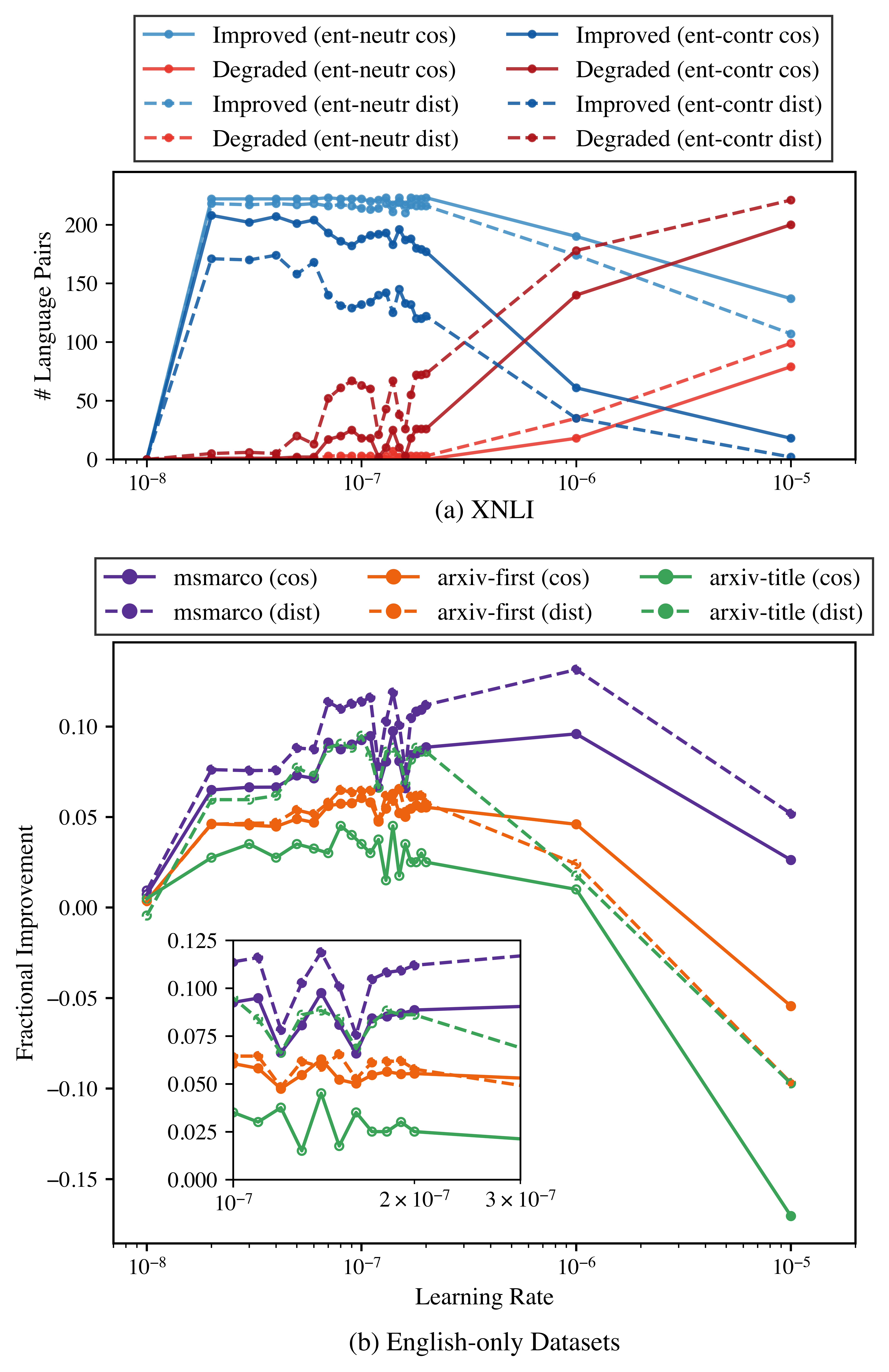}
    \caption{Evaluations on (a) XNLI and (b) the English-only datasets (MSMARCO and ARXIV) of the E5 query encoder tuned with a frozen embedding block, batch size 14, margin 0.1 using different learning rates. Values that did not pass the two-tailed test are shown with open markers.}
    \label{fig:e5_tuned_by_lr}
\end{figure}

Increasing the tuning learning rate delivers more gains on MSMARCO, while eventually reducing gains on XNLI and even ARXIV. Improvement of PND on MSMARCO and ARXIV is shown in Figure~\ref{fig:e5_tuned_by_lr}(b); the number of language pairs improved and degraded is in Figure~\ref{fig:e5_tuned_by_lr}(a). Appendix~\ref{app:learning_rate} contains the corresponding plots (Figure~\ref{fig:E5_L12_E5e_perf}) for the fully tuned $E5$ dual encoder, and for the $L12$ and $E5e$ models.
It is interesting that the $E5e$ model, not even being multilingual, still improves more than it degrades its rudimentary multilingual qualities. The effects of other tuning parameters are described in Appendix~\ref{app:tunings_parameters}. For example, the square-root batch size scaling rule works better than linear.

If we consider XNLI and ARXIV as indicators of how well a model keeps the learned skills while improving on narrow goals (e.g. MSMARCO), then our observation suggests there may be a slow tuning regime, at which the model preserves or even improves the existing skills which are at least a little related to the new goal. We call this \textit{adiabatic tuning}, in analogy to the slow process in quantum mechanics (a system starting in an eigenstate is kept in the same evolving eigenstate). For $E5$ the learning rates between 2e-8 and 6e-8 may be considered as the best. 

Our tentative explanation of adiabatic tuning is as follows: At low learning rates of tuning, the system (the encoder weights) remains in the 'minimum' region found at pretraining. This 'minimum' region is probably a wide well with uneven ground; the pretraining happened to terminate at some point inside the well. During tuning, the pretraining weight-space of twin encoder becomes just another surface in a family of surfaces, because of the added dimensions (the difference between the weights of the two encoders). We assume that due to continuity, the 'minimum' region, even if being reshaped, remains a well as the query encoder weights drift away from the weights of the text encoder. Within this well, improvements of all qualities related to the former, pretraining loss, may be still correlated. But if, at high learning rate, the model is strongly modified at some iteration (i.e. by backpropagation on a particular batch), then it may move away from the well.

\subsection{Extending adiabatic tuning range}\label{ssec:extend_adiabatic_range}

\begin{figure}[h!]
    \centering
    \includegraphics[width=\linewidth]{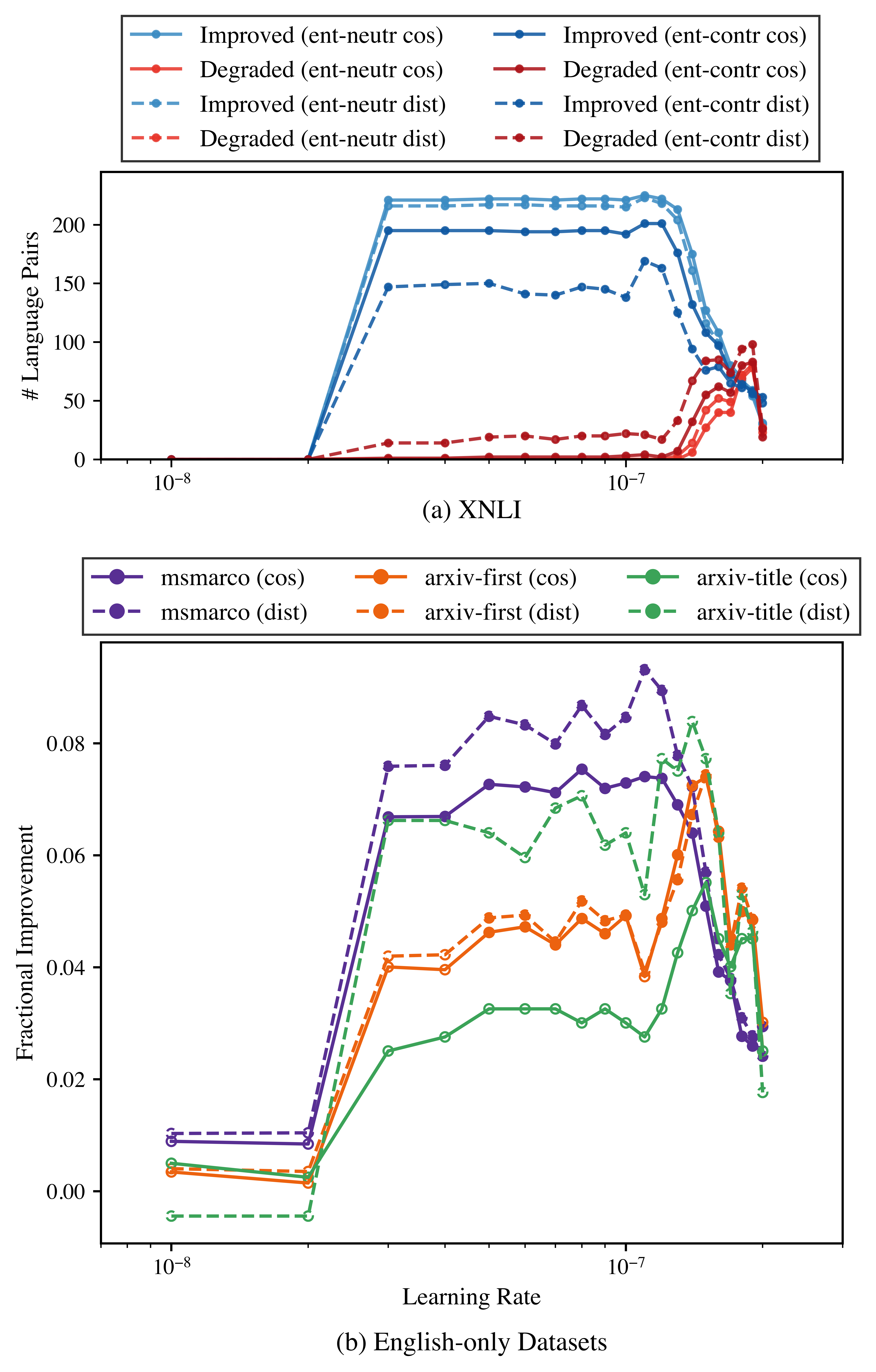}
    \caption{Evaluations of the $E5$ query encoder tuned with a frozen embedding block and all layers `out-put.dense.weight’, with batch size 14, margin 0.1 using different learning rates on (a) XNLI and (b) the English-only datasets (MSMARCO and ARXIV). Values that did not pass the two-tailed test are shown with open markers.}
    \label{fig:E5_frozen_dense}
\end{figure}

From evaluation results in Figure~\ref{fig:e5_tuned_by_lr} we may consider the learning rate below 7e-8 (but above 1e-8) as safely suitable for adiabatic tuning. But we know this only because we evaluated the tuned models on the out-of-tuning domains ARXIV and XNLI.

Is there any way to know the upper boundary without having extensive data for evaluation? Could there be an empirical recommendation not to exceed certain learning rate? Can we increase the adiabatic tuning range of learning rate?

In attempting to answer these questions, we have considered the largest changes in the layers at different learning rates. One suspect layer, by simple crude measures, is \textit{output.dense.weight}. In Appendix~\ref{app:freeze_suspects} in Tables~\ref{tab:layers_suspect} and~\ref{tab:layers_suspect_2} we show the most changing layers and the blocks to which they belong. Our motivation here is based on a simple and crude criteria; more detailed research and understanding may reveal better ways to extend the adiabatic tuning regime. 

The gains from the tuning by freezing the layer \textit{output.dense.weight} (in each transformer block) are shown in Figure~\ref{fig:E5_frozen_dense}. 
In comparison to the default tuning (Figure~\ref{fig:e5_tuned_by_lr}) we can see that the adiabatic regime indeed extends from a learning rate of about 6e-8 (as was in Figure~\ref{fig:e5_tuned_by_lr}) to about 1.3e-7. Thus, freezing of \textit{output.dense.weight} did help to somewhat extend the adiabatic tuning regime. However, this did not improve the gains, and further increase of the learning rate results in worse deterioration for the version with frozen \textit{output.dense.weight} layer, as can be seen for XNLI starting from the rate 1.4e-7. 

Another way of trying to stay longer in the original 'minimum' region during tuning could be by reducing the inertia of the optimizer. We present a simple attempt in Appendix~\ref{app:optim}, but the results are mixed.

\section{Conclusion}\label{sec:Conclusion}
We considered tuning the query part of a dual encoder starting from a high quality multilingual embedding model, and using English-only samples in the tuning. We found that multilingual qualities are quite stable in many scenarios of the tuning, and can be not only preserved but improved. We explain this by speculating that most of the transformer, except the embedding block, depends weakly on multiple languages.
We think of this as a particular case of a general pattern: tuning a certain model quality, if done carefully enough (\textit{adiabatic tuning}), can also retain or even improve the related (but not targeted by tuning) qualities.
This allows a resource-light adjustment of multilingual embeddings for a specific query type or domain, even a narrow domain (Appendix~\ref{app:narrow_tuning}).

\section*{Limitations}\label{sec:Limitations}
Our considerations here are limited to starting with a single high quality multilingual embedding model, and tuning it (on English-only samples) as a query encoder. While this setup is good for our understanding and convenient for adjusting an existing model, it would be natural to follow this up by considering a pre-trained multilingual dual encoder which is already asymmetric from the start. 

For our illustration we used the state of the art multilingual model \textit{intfloat/multilingual-e5-small}, and also, for comparison, repeated the same observations for the \textit{sentence-transformers/paraphrase-multilingual-MiniLM-L12-v2} model. We also repeated some of observations on monolingual model \textit{intfloat/e5-small-v2} - the tuning improved its rudimentary multilingual properties as well. Still, to gain a better understanding of the observed behaviors, it would be interesting to investigate more multilingual models.

We considered tuning the query encoder on English-only samples, and found that such tuning can ``pull up'' the quality of other languages too. 
Choosing another language for tuning would be interesting both for understanding and as a practical scenario.

We used MSMARCO triplets for tuning; we also verified some observations for models tuned on ARXIV-based subsets limited to a category (math, physics or cs, Appendix~\ref{app:narrow_tuning}). For evaluation we used a set aside part of MSMARCO triplets, and ARXIV in two variations, and XNLI. The motivation was that the MSMARCO evaluation part must show improvement (after tuning), ARXIV must verify the robustness of the improvement on a very different kind of texts (jargon-heavy), and XNLI must reveal the effect of the English-only driven improvement on multilingual qualities. We also confirmed the tuning gains on SQUAD and HotpotQA (both of which are quite different from MSMARCO). That said, the evaluations can be extended to even more datasets.

More research could be helpful in understanding and identifying the range of adiabatic tuning.


\bibliography{custom}

\appendix

\section{Usage of MSMARCO Triplets}\label{app:msmarco}
The MSMARCO dataset consists of 499184 samples, with each sample being a tuple given as (query, positives, negatives).  The ``positives'' are the correct answers to the query, and the ``negatives'' are semantically similar, but incorrect answers.  For most samples, there is only one positive, but many negatives. For our tuning we simply select the very first positive and the very first negative. 
Thus, each sample gives one triplet (anchor, positive, negative) for contrastive learning, where the query is taken as an anchor. 

We keep the first 487983 samples (or 34856 batches if each batch is 14 triplets) for tuning, leaving the next 4200 samples (300 batches) for validation, and the last 7000 samples for evaluation. During evaluation we create all possible triplets from the 7000 samples, using all positives and negatives; this makes 357642 evaluation triplets.

Almost half of MSMARCO samples have the maximal number of negatives (65), and for evaluation shown in Table~\ref{tab:improve_E5default_ondatasets} we use a more difficult version 'MSMARCO 65 negatives', with all samples with less than 65 negatives filtered out.

\section{ARXIV Dataset for Triplets}\label{app:arxiv}
\subsection{Dataset arxiv-negatives}\label{sapp:dataset_arxiv}

Of ARXIV we use titles and abstracts. 
In order to have a representative subset of a manageable size for our evaluations, we select all samples that have at least one category with a maximum size of 10K samples. For example, the arxiv category \textit{bayes-an} is the smallest (size 16) in our snapshot (version 173), meaning that there were only 16 arxiv preprints in this category. 

We made two flavors of evaluation arxiv triplets from this arxiv subset. In the first version, the anchor is the title, the positive is the corresponding abstract, and the negative is another random abstract. In the second version the anchor is the first sentence of the 'positive' abstract, the positive is the rest of the abstract, and the negative is a similar piece (first sentence excluded) of the 'negative' abstract.

We make use of triplets created from arxiv because this provides our evaluation with a very different kind of text (compared to MSMARCO), and thus allows us to judge the robustness of the improvement.
For convenience and reproducibility of creating triplets of different levels of difficulty, we made a dataset \textit{arxiv-negatives}\footnote{https://huggingface.co/datasets/primer-ai/arxiv-negatives}.

The dataset consists of 253140 samples, each sample is a tuple of two elements:
\begin{enumerate}[topsep=0pt,itemsep=-1ex,partopsep=1ex,parsep=1ex]
    \item An ARXIV paper metadata, including its Id, title and abstract and categories.
    \item List of 21 Ids of other ARXIV papers. The first 20 Ids are the papers that are 'closest' to the above paper, and sorted from the most to the least similar; the last 21st Id is an Id of a randomly selected paper (not coinciding with Id of the above paper).
\end{enumerate}
Thus, we have 21 versions of picking up negatives for triplets, from the most difficult to the easiest (the last one, of the random selection).

For example, to create triplets of difficulty 14, for each paper given by the first tuple element, we pick up a paper corresponding to 14th Id given in the second tuple element. From the first paper we can create query and positive, and from the second paper, negative. Through this work we used two flavors:
\begin{enumerate}[topsep=0pt,itemsep=-1ex,partopsep=1ex,parsep=1ex]
    \item `Title': The title of the first paper acts as the query and its abstract as the positive; the negative is then the abstract of the second paper.
    \item `First': The query is the first sentence of the abstract of the first paper; the positive is the rest of the abstract; the negative is the abstract of the second paper, with its first sentence deleted.
\end{enumerate}
\subsection{How is it created?}\label{sapp:arxiv_how_created}
The above dataset is created from the mirror of arxiv (version 173) \textit{arxiv-metadata-oai-snapshot.jsonl} through the following steps:
\begin{enumerate}[topsep=0pt,itemsep=-1ex,partopsep=1ex,parsep=1ex]
    \item Identified all arxiv categories with a maximum size of 10K papers (i.e. arxiv preprints).
    \item Selected all papers that have at least one of the categories identified above. This is the subset of arxiv to deal with: manageably small, yet diverse.
    \item For each paper: (1) Sort its categories by size, from smaller to larger. (2) Find all other papers that have the closest match by the categories (the closest match is the longest consecutive list of matched categories, starting from the first one). (3) Of the found papers, select 20 closest by Jensen-Shannon distance between the paragraphs, and sort them by the distance. If there were less than 20 papers, fill to 20 by the last one. (4) Add randomly selected paper as 21st.
\end{enumerate}
Of the total 253140 samples, in 213156 samples (84.2\%) all the first 20 negatives are different (which means that not less than 20 papers happen to have the same closest match by categories).

\section{SQUAD}\label{app:dataset_squad}
For using the SQUAD dataset, we identified (for each query) the given paragraph sentences containing an answer to the query as positives, and the rest of the sentences as negatives. We left samples having at least 1 positive and 1 negative. On average there is 1.3 positives and 4.2 negatives per a query. For the evaluation shown in Table~\ref{tab:improve_E5default_ondatasets} we combined train, validation and test subsets. The results are given also for a version called `SQUAD min 5', in which we have filtered out queries that had less than 5 candidate sentences.

\section{HotpotQA}\label{app:dataset_hotpotqa}
For using HotpotQA, we combined its train and dev subsets. For each query (`question') both train and dev subsets contain on average 9.95 passages, of which 2 are always positives. For the evaluation shown in Table~\ref{tab:improve_E5default_ondatasets} we filtered out queries that had less than 10 passages, and split the dataset into `easy', `medium' and `hard' subsets accordingly to the HotpotQA labels of the difficulty of the samples. 

\section{Tuning}\label{app:tuning}
Unless specified otherwise, we tune a dual encoder by contrastive learning in the following simple regime: 
\begin{enumerate}[topsep=0pt,itemsep=-1ex,partopsep=1ex,parsep=1ex]
    \item The text encoder is fully frozen; the frozen parts of the query encoder are specified. 
    \item The batch size is 14, the learning rate is \mbox{5e-8} and the contrastive learning margin is $0.1$. The loss is defined by the triple margin loss.
    \item There are $1000$ batches per epoch, i.e. 14000 samples per epoch.
    \item Stopping occurs after $10$ consecutive non-improvement epochs. The improvement is measured on the validation subset after each epoch.
    The model is considered to be improved if (on the validation subset) both the loss and the count of errors have decreased.
    \item The AdamW optimizer is used.
\end{enumerate}
Changing this default regime is considered in Appendixes~\ref{app:learning_rate},~\ref{app:tunings_parameters}.

\section{XNLI}\label{app:xnli}
The XNLI dataset consists of pairs of sentences which are human-labeled as entailment, neutral or contradiction. The test subset (which we use) contains 1670 pairs for each of these labels and each sentence is presented in 15 languages: ['ar', 'bg', 'de', 'el', 'en', 'es', 'fr', 'hi', 'ru', 'sw', 'th', 'tr', 'ur', 'vi', 'zh']. We use 225 versions of the pairs, because each sentence of the pair can be in any of the 15 languages. At evaluation the first sentence serves as the query (the embedding is taken by the query model), and the second one as the text. We expect that the sentences of an entailment pair should be closer to each other than the sentences of any neutral pair, or of any contradiction pair. Whenever this does not happen, we count this as an error.

\section{Performance of Untuned Query Encoder}\label{app:performance_untuned}
To establish a baseline before any fine-tuning, and to ensure our evaluation is not too easy, we measure the errors of the original $E5$ model on the data described in Section~\ref{ssec:Tuning} and show the results in Table~\ref{tab:PND_untuned}. We also measure the errors of $L12$ and of $E5e$ - a more recent monolingual (English) model.

\begin{table}[th!]
\centering
\begin{tabular}{@{}clrrr@{}}
\hline
{data}&{Evaluation}&{$E5$}&{$L12$}&{$E5e$}\\
\hline
    \parbox[t]{2mm}{\multirow{3}{*}{\rotatebox[origin=c]{90}{MM}}}&{N tot}&\multicolumn{2}{c}{357642}\\
    {}&PND (cos)&{4.7\%}&{15.1\%}&{4.6\%}\\
    {}&PND (dist)&{4.8\%}&{15.4\%}&{4.5\%}\\
\hline
    \parbox[t]{2mm}{\multirow{3}{*}{\rotatebox[origin=c]{90}{ARX-F}}}&{N tot}&\multicolumn{2}{c}{253140}\\
    {}&PND (cos)&{1.6\%}&{4.9\%}&{3.1\%}\\
    {}&PND (dist)&{1.6\%}&{6.7\%}&{3.5\%}\\
\hline
    \parbox[t]{2mm}{\multirow{3}{*}{\rotatebox[origin=c]{90}{ARX-T}}}&{N tot}&\multicolumn{2}{c}{253140}\\
    {}&PND (cos)&{0.2\%}&{1.4\%}&{0.2\%}\\
    {}&PND (dist)&{0.2\%}&{1.7\%}&{0.2\%}\\
\hline
    \parbox[t]{2mm}{\multirow{5}{*}{\rotatebox[origin=c]{90}{XNLI}}}&{N total}&\multicolumn{2}{c}{2788900}\\
    {}&PND e-n (cos)&{10.8\%}&{10.2\%}&{15.9\%}\\
    {}&PND e-c (cos)&{10.0\%}&{7.2\%}&{15.3\%}\\
    {}&PND e-n (dist)&{10.5\%}&{10.1\%}&{15.9\%}\\
    {}&PND e-c (dist)&{9.6\%}&{7.8\%}&{15.4\%}\\
\hline
\end{tabular}
\caption{The count of errors for the original untuned models $E5$, $L12$ and $E5e$, on the datasets noted in the first column: \textit{MM} - MSMARCO test 7000 samples (357642 triplets, see Section~\ref{ssec:Datasets} and Appendix~\ref{app:msmarco}); \textit{ARX-F} - arxiv-first, the arxiv subset with the abstract's first sentence as an anchor; \textit{ARX-T} - arxiv-title, the arxiv subset with the title as an anchor; \textit{XNLI} - XNLI test subset providing 1670x1670=2788900 comparisons of entailment pairs vs neutral pairs (and the same amount of entailment pairs vs contradiction pairs). For XNLI the errors are averaged over 225 (15x15) language-language versions, and shown as percent of $N total$. The evaluation is done using cosine similarity or euclidean distance similarity (\textit{cos} or \textit{dist} in second column).}
\label{tab:PND_untuned}
\end{table}

\begin{figure*}[h!]
    \centering
    \includegraphics[width=\linewidth]{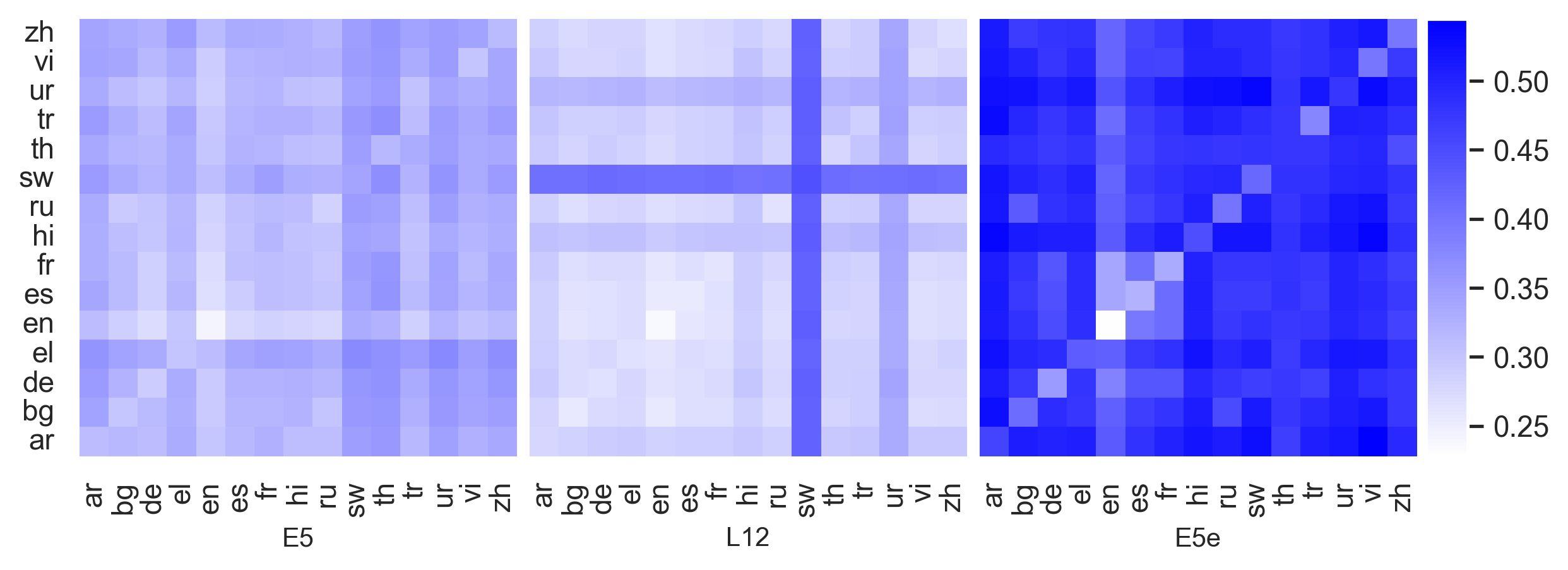}
    \caption{PND of embedding models on XNLI entailment-neutral comparisons assessed by cosine.}
    \label{fig:hmap_xnli_orig_c_10}
\end{figure*}

\begin{figure*}[h!]
    \centering
    \includegraphics[width=\linewidth]{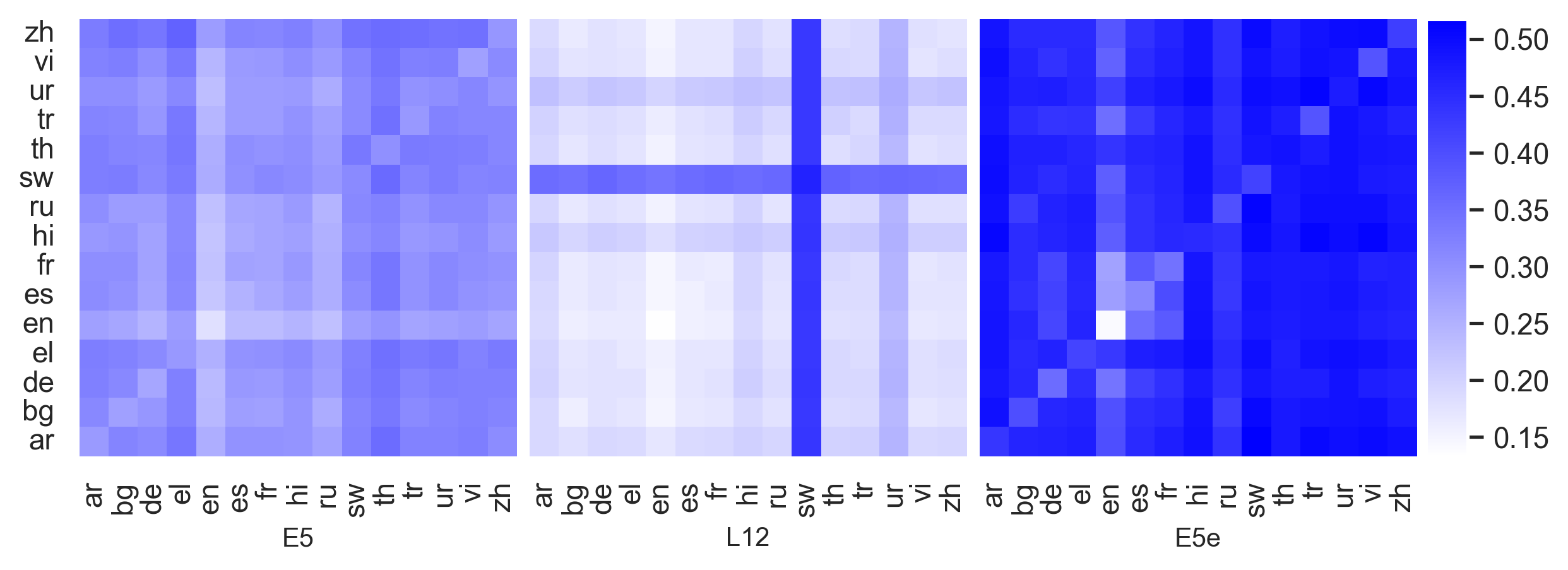}
    \caption{PND of embedding models on XNLI entailment-contradiction comparisons assessed by cosine.}
    \label{fig:hmap_xnli_orig_c_20}
\end{figure*}

The count of errors on the triplets (MSMARCO, ARXIV) is straightforward: it is an error when a positive is not closer than a negative to the anchor of the triplet. On XNLI we sum up the error count over all language-language pairs and divide the sum by the number (255=15x15) of such pairs. This averaged error is shown as a percentage of the total (2788900) comparisons; each comparison here is either a comparison of an entailment-labeled sample with a neutral-labeled sample (\textit{entail-neutral} in the table) or a comparison of an entailment-labeled sample with a contradiction-labeled sample (\textit{entail-contr} in the table). An error was counted whenever the sentences of an entailment sample happened to be farther from each other than the sentences of a neutral (or contradiction) sample. Separately for each pair of languages PND is shown in Figures~\ref{fig:hmap_xnli_orig_c_10},~\ref{fig:hmap_xnli_orig_c_20} for cosine similarity measure. The distance measure gives results visually almost undistinguishable.

The amount of errors in Table~\ref{tab:PND_untuned} and in Figures~\ref{fig:hmap_xnli_orig_c_10},~\ref{fig:hmap_xnli_orig_c_20} is reasonable enough to consider how tuning would affect them. The smallest counts are the counts of positive-negative discrepancies of $E5$ and $E5e$ on ARX-T (apparently, a title makes an easier 'query' than the first sentence of an abstract). These counts are 309 and 420 for the cosine similarity (the row ARX-T PND (cos)), and 453 and 580 for the distance similarity (the row ARX-T PND (dist)).

Notice that $L12$ has far worse PND on English data (MSMARCO and ARXIV). The English-only model $E5e$, as expected, performs worse than multilingual models $E5$ and $L12$ on multilingual XNLI, but its PND is still far below $50\%$, because there is much similarity between some of the languages.

\section{Gains on ARXIV for Different Levels of Difficulty}\label{app:arxiv_difficulty_results}

\begin{figure}[h!]
    \centering
    \includegraphics[width=\linewidth]{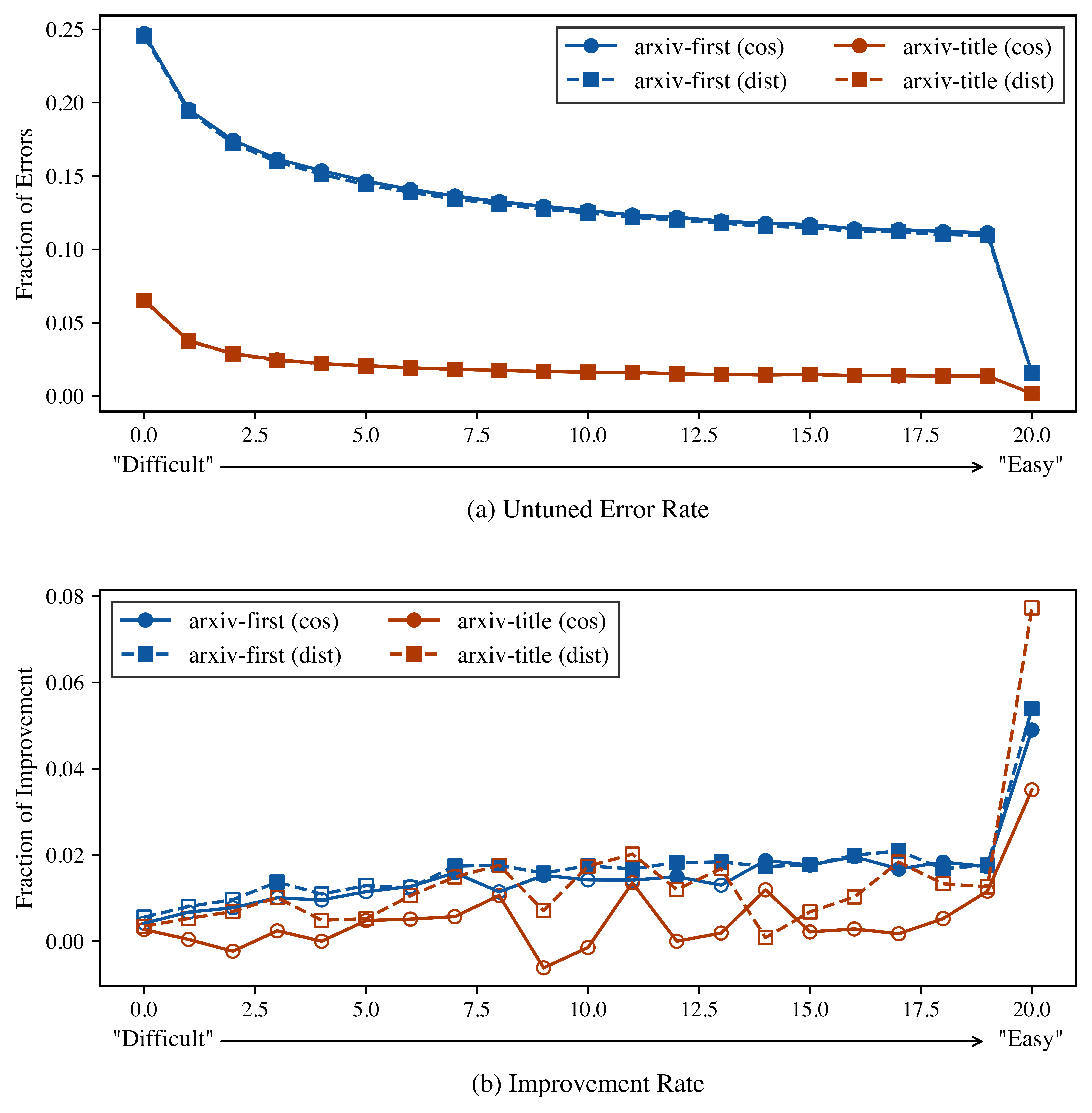}
    \caption{Errors and improvements on arxiv-negatives dataset of different level of difficulty. The ``easiest'' dataset is a random selection of negatives from the same data used through this work in evaluations. In (a), we show the fraction of errors done by the original E5 model (for comparison, see Table~\ref{tab:PND_untuned}).  In (b), we show the improvement after tuning the query encoder on MSMARCO, with `default’ settings, i.e. learning rate 5e-8, batch size 14, margin 0.1 and frozen embedding block. Values that did not pass the two-tailed test (Appendix I) are shown with open markers.}
    \label{fig:arxiv_neg_perf}
\end{figure}

Throughout the paper we used the easiest version of triplets in the arxiv-negatives dataset, the version that uses randomly selected negatives. Here in Figure~\ref{fig:arxiv_neg_perf} we show, for comparison, the fraction of the errors which occur in the original untuned $E5$ embeddings using the other levels of difficulty, and also the corresponding improvements (by Equation~\ref{eq:improvement}) after tuning the query encoder on the MSMARCO with frozen embedding block and our default settings (Section~\ref{ssec:Tuning}). The statistical significance of the improvements in Figure~\ref{fig:arxiv_neg_perf} is estimated as explained in Appendix~\ref{app:significance}.

The difficulty of intentionally close negatives is much harder, but Figure~\ref{fig:arxiv_neg_perf} still shows that performance on ARXIV was mostly improved. We used the easiest triplets version for our evaluations throughout the paper because it more distinctly indicated the trends in the improvements.


\section{Significance Test}\label{app:significance}
In Table~\ref{tab:improve_E5_freeze}, Figure~\ref{fig:e5_tuned_by_lr} and through the paper we use two-proportion $Z$-test, pooled for $H_0:p_1=p_2$. We are comparing the number of errors original $n_0$ and improved $n_1$, having the total $N$ (the totals can be seen in Table~\ref{tab:PND_untuned}); a total is the same for original and improved version. We deem the difference to be significant if $|Z|>Z_c$ where
\begin{equation} \label{eq:significance}
Z = \frac{p_1-p_0}{\sqrt{\frac{1}{2}P(1-P)N}}
\end{equation}
with $p_0=n_0/N$, $p_1=n_1/N$ and $P=\frac{1}{2}(n_0+n_1)/N$. We used $Z_c=1.96$, which is a critical value corresponding to probability 0.975.

Notice that in our examples the values $N$ are typically very large. And the improvements we report, according to Equation~\ref{eq:improvement}, are relative, not absolute values.

\section{Encoder L12 with Frozen Layers}\label{app:L12_freezing}
Table~\ref{tab:improve_L12_freeze} shows results of tuning with freezing some of $L12$ layers. It is similar to the Table~\ref{tab:improve_E5_freeze} for $E5$. And, similar to $E5$, freezing everything except the embedding, resulted in negligible changes of the query encoder (not shown in the table).

\begin{figure*}[h!]
    \centering
    \includegraphics[width=\linewidth]{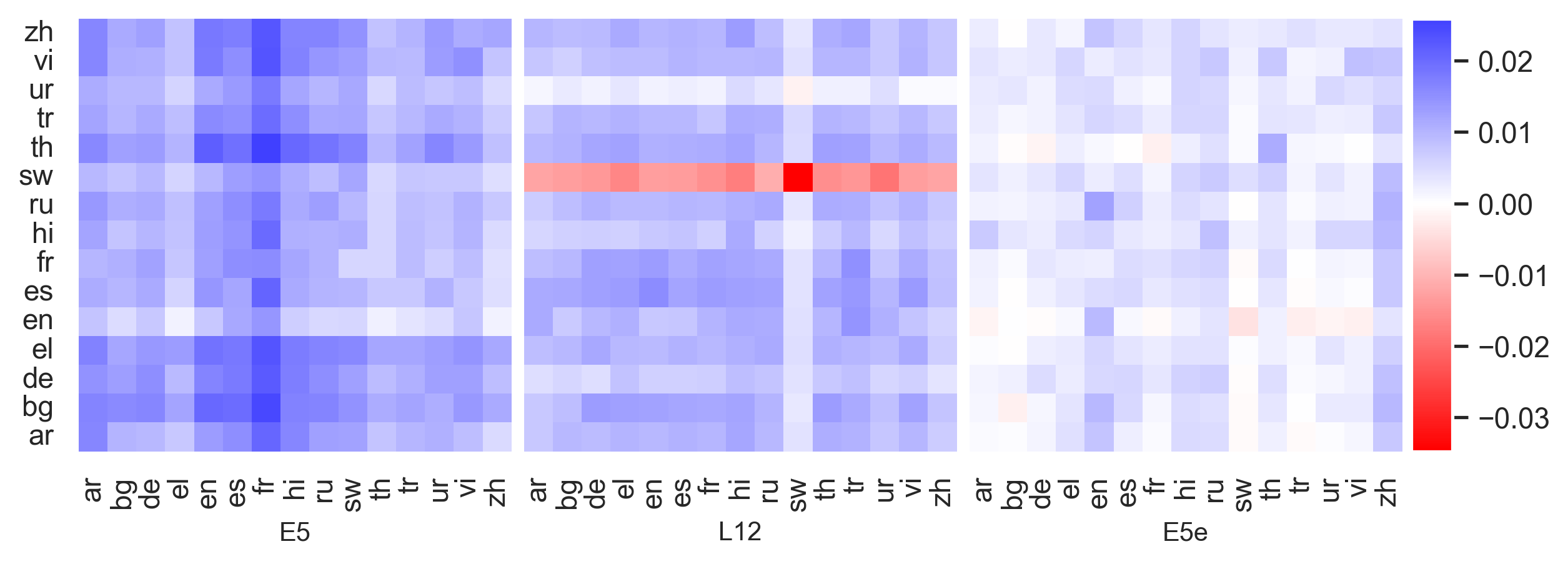}
    \caption{Improvement of $E5$, $L12$ and $E5e$ on XNLI entailment-neutral comparisons assessed by cosine.}
    \label{fig:hmap_xnli_gain_c_10}
\end{figure*}

\begin{figure*}[h!]
    \centering
    \includegraphics[width=\linewidth]{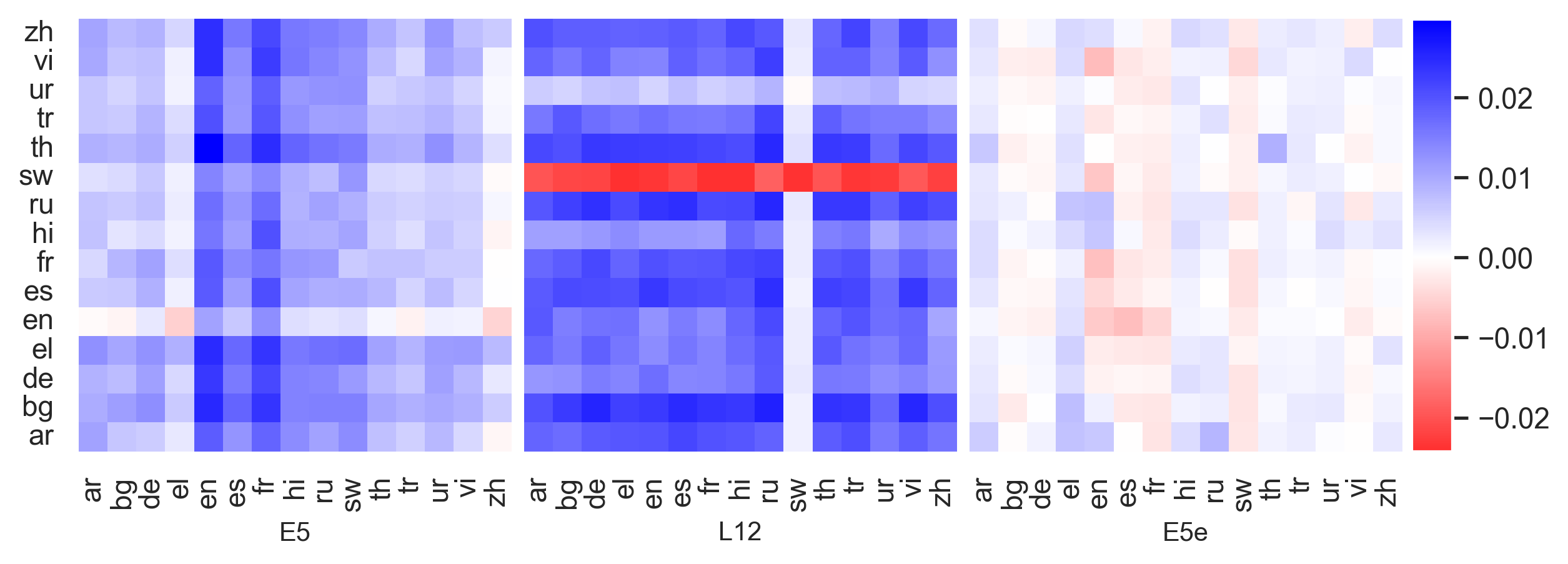}
    \caption{Improvement of $E5$, $L12$ and $E5e$ on XNLI entailment-contradiction comparisons assessed by cosine.}
    \label{fig:hmap_xnli_gain_c_20}
\end{figure*}

The changes in cross-lingual qualities corresponding to the third row (\textit{emb}, frozen embedding block) of Table~\ref{tab:improve_L12_freeze} are shown in comparison with $E5$ and $E5e$ embeddings in Figures~\ref{fig:hmap_xnli_gain_c_10} and~\ref{fig:hmap_xnli_gain_c_20}. Note that $E5e$ is not a multilingual embedding model. Having a worse start as a multilingual embedding model, $E5e$ also gets much weaker improvements of its multilingual qualities; it is consistent with our understanding of adiabatic tunings (Section~\ref{ssec:adiabatic tuning}).

\begin{table*}[th!]
\centering
\begin{tabular}{@{}l|rr|rr|rr|ll|ll@{}}
\toprule
{} & \multicolumn{2}{c|}{msmarco} & \multicolumn{2}{c|}{arxiv-first} & \multicolumn{2}{c|}{arxiv-title} & \multicolumn{2}{c|}{xnli ent-neutr} & \multicolumn{2}{c}{xnli ent-contr}\\
{frozen}&{c\%}&{d\%}&{c\%}&{d\%}&{c\%}&{d\%}&{c+/-}&{d+/-}&{c+/-}&{d+/-}\\
\hline
    {-}&{6.60}&{6.93}&{2.65}&{-7.86}&{14.46}&{\cellcolor{grey!10}{-0.12}}&{206/15}&{57/35}&{201/15}&{15/189}\\
    {emb.base}&{7.28}&{8.04}&{2.46}&{-9.17}&{12.38}&{\cellcolor{grey!10}{-1.03}}&{200/15}&{47/47}&{206/15}&{20/142}\\
    {emb}&{7.28}&{8.04}&{2.51}&{-9.17}&{12.4}&{\cellcolor{grey!10}{-1.03}}&{200/15}&{47/47}&{206/15}&{20/143}\\
    {emb, B0a}&{7.26}&{8.03}&{\cellcolor{grey!10}{2.29}}&{-9.22}&{12.52}&{\cellcolor{grey!10}{-0.96}}&{201/15}&{46/46}&{206/15}&{20/142}\\
    {emb, B0a,i}&{7.03}&{7.75}&{2.44}&{-8.6}&{12.46}&{\cellcolor{grey!10}{-0.63}}&{203/15}&{51/41}&{206/15}&{20/133}\\
    {emb, B0a,i,od}&{9.04}&{10.15}&{\cellcolor{grey!10}{1.80}}&{-17.54}&{12.49}&{-8.58}&{195/19}&{30/116}&{207/15}&{19/167}\\
    {emb, B0}&{8.92}&{9.98}&{\cellcolor{grey!10}{1.78}}&{-16.73}&{12.88}&{-8.07}&{195/16}&{33/102}&{209/15}&{20/163}\\
    {emb, B0-5}&{8.54}&{9.68}&{2.71}&{-19.01}&{12.35}&{-12.17}&{209/15}&{28/129}&{209/15}&{19/172}\\
    {emb, B0-10}&{\cellcolor{grey!10}{0.11}}&{\cellcolor{grey!10}{0.15}}&{\cellcolor{grey!10}{0.10}}&{\cellcolor{grey!10}{-0.12}}&{\cellcolor{grey!10}{0.25}}&{\cellcolor{grey!10}{-0.02}}&{0/0}&{0/0}&{0/0}&{0/0}\\
\hline
\end{tabular}
\caption{Evaluations of the $L12$ query model tuned on MSMARCO as described in Section~\ref{ssec:Tuning}.
The notations are as in Table~\ref{tab:improve_E5_freeze}.}
\label{tab:improve_L12_freeze}
\end{table*}

\section{Narrow-Domain Query Encoder}\label{app:narrow_tuning}
\begin{figure*}[h!]
    \centering
    \includegraphics[width=\linewidth]{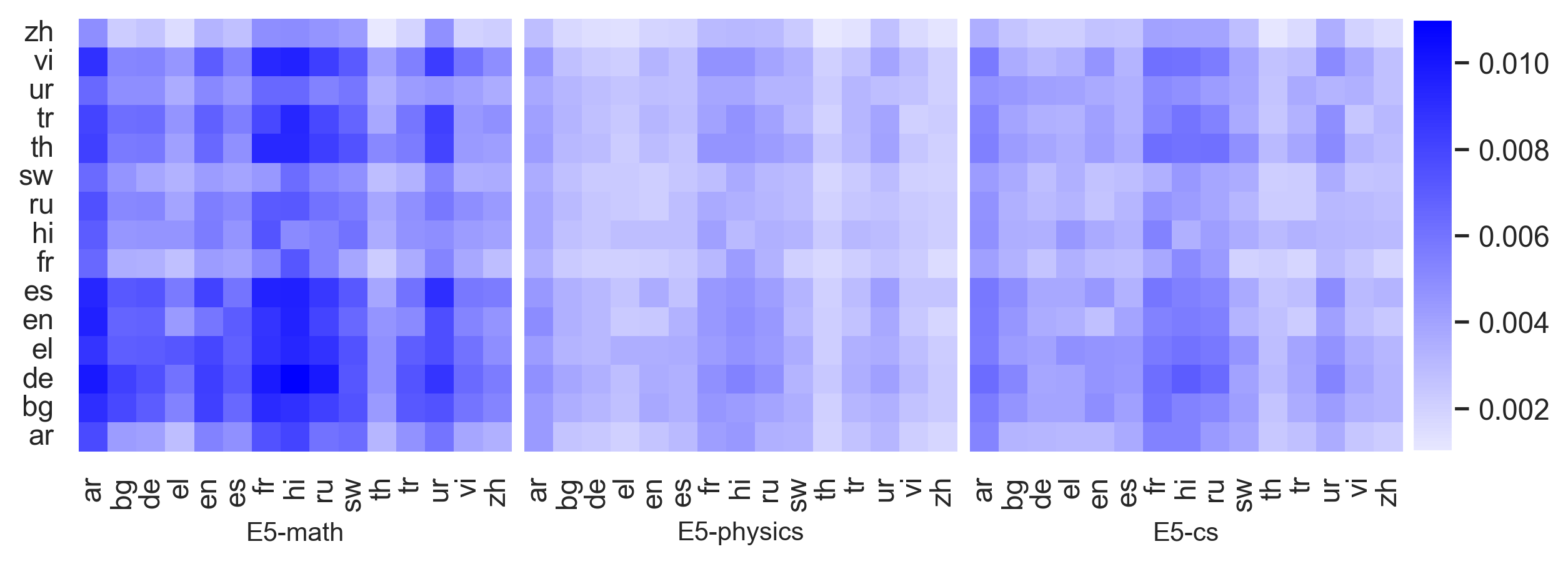}
    \caption{Improvement of narrow-tuned encoders on XNLI entailment-neutral comparisons assessed by cosine.}
    \label{fig:hmap_xnli_gain_narrow_c_10}
\end{figure*}

\begin{figure*}[h!]
    \centering
    \includegraphics[width=\linewidth]{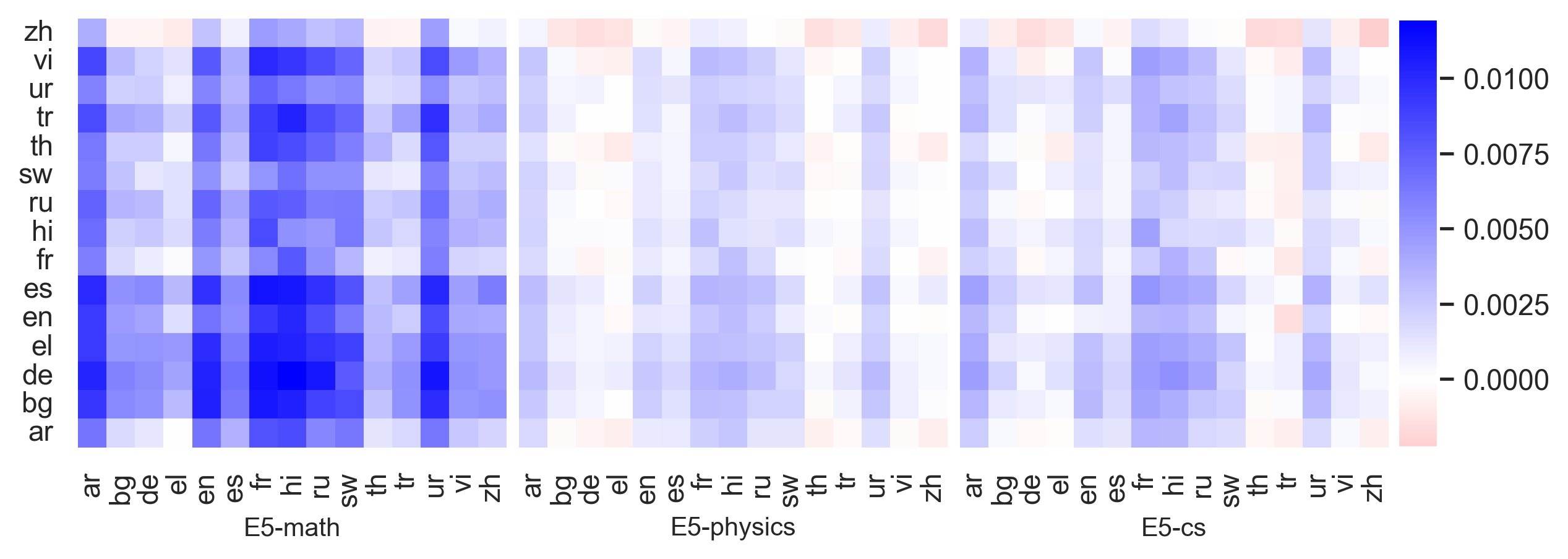}
    \caption{Improvement of narrow-tuned encoders on XNLI entailment-contradiction comparisons assessed by cosine.}
    \label{fig:hmap_xnli_gain_narrow_c_20}
\end{figure*}

So far we observed that tuning the query encoder on data of a certain style (MSMARCO dataset) could preserve (or even improve) the encoder qualities which are not targeted by the tuning task, especially if we tune with a frozen embedding layer and low learning rate. Here we provide observations using more specialized datasets, based on arxiv-first (arxiv-first is described in Section~\ref{ssec:Datasets} and Appendix \ref{app:arxiv}):
\begin{enumerate}[topsep=0pt,itemsep=-1ex,partopsep=1ex,parsep=1ex]
    \item ARXIV-math: uses only documents with at least one category which has the prefix "math."
    \item ARXIV-physics: As above, but with "physics." as the prefix
    \item ARXIV-cs: As above, but with "cs." as the prefix 
\end{enumerate}
$E5$ tuned on these narrow datasets using our `default' regime (Section~\ref{ssec:Tuning}) with frozen embedding block mostly improves the PND (positive-negatives discrepancy fraction) as shown in Table~\ref{tab:Narrow_models_PND}.
The improvements of these narrow-tuned encoders on individual language pairs, assessed by cosine, are shown in Figures~\ref{fig:hmap_xnli_gain_narrow_c_10} and \ref{fig:hmap_xnli_gain_narrow_c_20}.

\begin{table*}[th!]
\centering
\begin{tabular}{@{}c|rr|rr|rr|ll|ll@{}}
\toprule
{} & \multicolumn{2}{c|}{msmarco} & \multicolumn{2}{c|}{arxiv-first} & \multicolumn{2}{c|}{arxiv-title} & \multicolumn{2}{c|}{xnli ent-neutr} & \multicolumn{2}{c}{xnli ent-contr}\\
{model}&{c\%}&{d\%}&{c\%}&{d\%}&{c\%}&{d\%}&{c+/-}&{d+/-}&{c+/-}&{d+/-}\\
\hline
    {E5-math}&{\cellcolor{grey!10}{0.04}}&{\cellcolor{grey!10}{0.45}}&{54.71}&{52.85}&{50.38}&{53.64}&{218/0}&{209/0}&{177/0}&{133/0}\\
    
    {E5-physics}&{\cellcolor{grey!10}{0.16}}&{\cellcolor{grey!10}{0.57}}&{19.05}&{18.64}&{\cellcolor{grey!10}{21.8}}&{\cellcolor{grey!10}{20.97}}&{162/0}&{102/0}&{31/0}&{2/0}\\

    {E5-cs}&{\cellcolor{grey!10}{0.18}}&{\cellcolor{grey!10}{0.55}}&{23.32}&{23.63}&{25.56}&{26.49}&{205/0}&{136/0}&{51/0}&{8/8}\\
    
\hline
\end{tabular}
\caption{Evaluations of the $E5$ query encoder tuned on ARXIV-math, ARXIV-physics or ARXIV-cs with a frozen embedding block, batch size 14, margin 0.1 and learning rate 5e-8. When evaluated on ARXIV (columns arxiv-first and arxiv-title) the samples with category of the model (the first column) are excluded from the evaluation data.}
\label{tab:Narrow_models_PND}
\end{table*}

\section{Learning Rate}\label{app:learning_rate}
\begin{figure*}[h!]
    \centering
    \includegraphics[width=\linewidth]{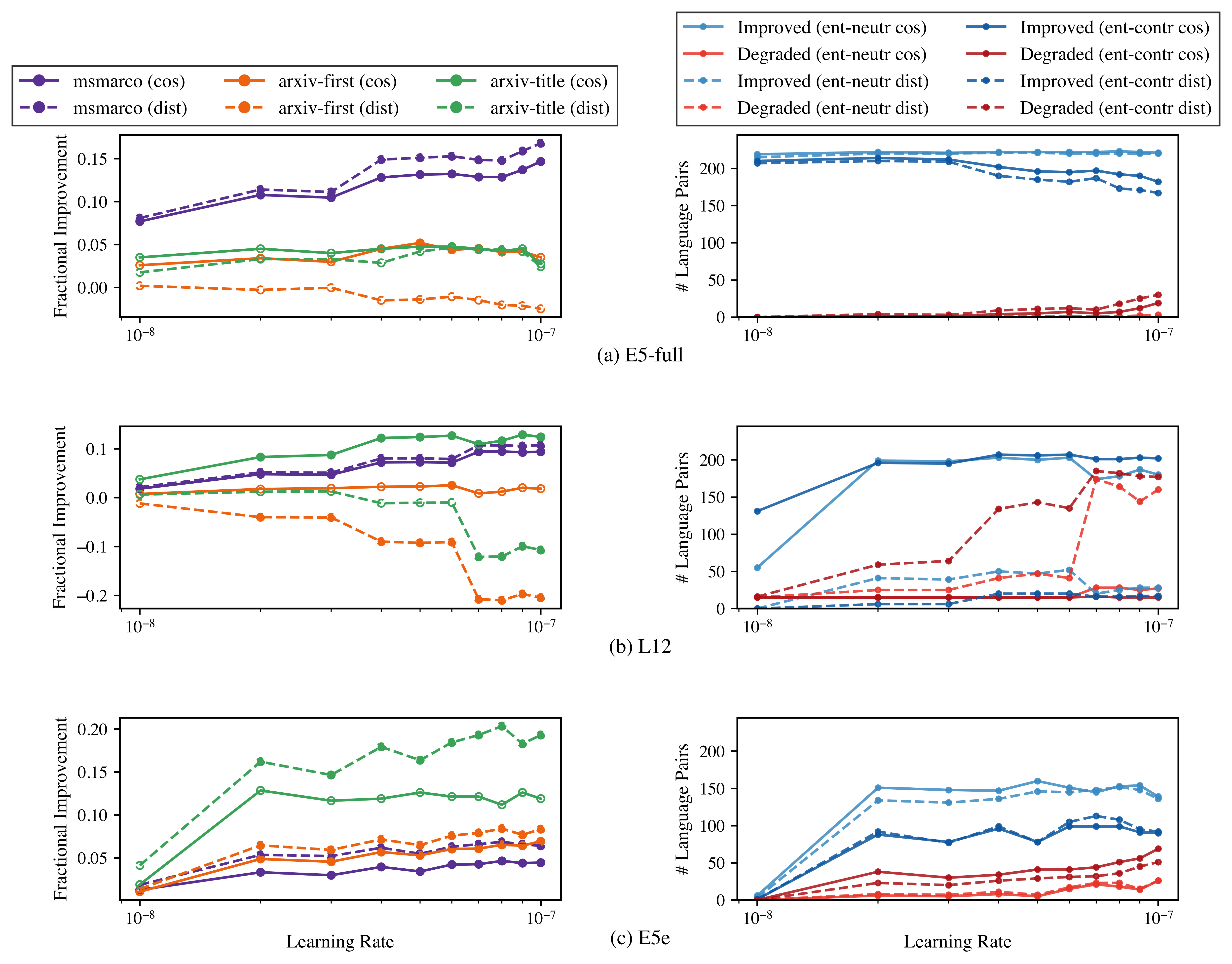}
    \caption{Improvement of various models and tuning configurations on the English-only datasets (MSMARCO and ARXIV) in the left column and XNLI in the right column.  Values that did not pass the two-tailed test (Appendix~\ref{app:significance}) are shown with open markers.  (a) Evaluations of the $E5$-full dual encoder after both encoders were tuned with a frozen embedding block, batch size 14 and margin 0.1.  (b) Evaluations of the $L12$ query encoder tuned with a frozen embedding block, batch size 14 and margin 0.1.  (c) Evaluations of the $E5e$ query encoder tuned with a frozen embedding block, batch size 14 and margin 0.1.}
    \label{fig:E5_L12_E5e_perf}
\end{figure*}

In Figure~\ref{fig:e5_tuned_by_lr} we have shown how the improvements of the $E5$ model depend on the learning rate. Here in Figure~\ref{fig:E5_L12_E5e_perf} we compare similar data for $L12$ and $E5e$ as well as a particular instance of $E5$ when both the query and text encoder are subject to tuning (as two independent encoders, with the same starting point) with the embedding block frozen in both encoders. The data confirm that while higher learning rates are not yet overtuning and still give higher gains on the test subset (of MSMARCO), it is the lower learning rates that better preserve and even improve those pretrained qualities which are not the goal of tuning.

\section{Tuning Regime}\label{app:tunings_parameters}

\subsection{Learning rate and batch size}\label{sec:scaling_rule}
\subsubsection{Scaling rule}\label{ssec:scaling_rule}

\begin{table*}[th!]
\centering
\begin{tabular}{@{}cc|rr|rr|rr|ll|ll@{}}
\toprule
{batch} & {learning} & \multicolumn{2}{c|}{msmarco} & \multicolumn{2}{c|}{arxiv-first} & \multicolumn{2}{c|}{arxiv-title} & \multicolumn{2}{c|}{xnli ent-neutr} & \multicolumn{2}{c}{xnli ent-contr}\\
{size}&{rate} & {c\%}&{d\%}&{c\%}&{d\%}&{c\%}&{d\%}&{c+/-}&{d+/-}&{c+/-}&{d+/-}\\
\hline
    {7}& {2.5e-8}&{6.62}&{7.67}&{4.48}&{5.06}&{\cellcolor{grey!10}{3.26}}&{\cellcolor{grey!10}{6.18}}&{221/0}&{216/0}&{201/2}&{160/13}\\
    
    {14}& {5.0e-8}&{7.30}&{8.82}&{4.90}&{5.39}&{\cellcolor{grey!10}{3.51}}&{\cellcolor{grey!10}{7.73}}&{222/0}&{217/1}&{201/2}&{158/20}\\
    
    {28}& {1.0e-7}&{8.36}&{10.31}&{5.71}&{6.75}&{\cellcolor{grey!10}{3.26}}&{\cellcolor{grey!10}{7.95}}&{222/0}&{218/3}&{177/20}&{128/58}\\
    
    {56}& {2.0e-7}&{8.32}&{10.54}&{5.39}&{6.75}&{\cellcolor{grey!10}{2.76}}&{\cellcolor{grey!10}{7.51}}&{222/0}&{217/3}&{193/14}&{141/42}\\
    
    {112}& {4.0e-7}&{8.46}&{10.36}&{5.24}&{6.00}&{\cellcolor{grey!10}{3.26}}&{\cellcolor{grey!10}{8.39}}&{221/0}&{216/3}&{197/8}&{147/35}\\
\hline
\end{tabular}
\caption{Evaluations of the $E5$ query encoder tuned with a frozen embedding block, margin 0.1 and 14000 samples per epoch. Linear scaling rule of learning rate with batch size. Values that did not pass the two-tailed test are shown in gray.}
\label{tab:Dependency_E5_batch_linear}
\end{table*}

\begin{table*}[th!]
\centering
\begin{tabular}{@{}cc|rr|rr|rr|ll|ll@{}}
\toprule
{batch} & {learning} & \multicolumn{2}{c|}{msmarco} & \multicolumn{2}{c|}{arxiv-first} & \multicolumn{2}{c|}{arxiv-title} & \multicolumn{2}{c|}{xnli ent-neutr} & \multicolumn{2}{c}{xnli ent-contr}\\
{size}&{rate} & {c\%}&{d\%}&{c\%}&{d\%}&{c\%}&{d\%}&{c+/-}&{d+/-}&{c+/-}&{d+/-}\\
\hline
    {7}& {3.54e-8}&{6.80}&{7.84}&{4.43}&{4.76}&{\cellcolor{grey!10}{2.26}}&{\cellcolor{grey!10}{6.18}}&{221/0}&{216/0}&{192/2}&{148/17}\\
    
    {14}& {5.00e-8}&{7.30}&{8.82}&{4.90}&{5.39}&{\cellcolor{grey!10}{3.51}}&{\cellcolor{grey!10}{7.73}}&{222/0}&{217/1}&{201/2}&{158/20}\\
    
    {28}& {7.07e-8}&{7.16}&{9.10}&{4.70}&{5.57}&{\cellcolor{grey!10}{4.01}}&{\cellcolor{grey!10}{7.95}}&{222/0}&{217/1}&{192/4}&{140/32}\\
    
    {56}& {1.00e-7}&{7.32}&{8.56}&{4.70}&{5.16}&{\cellcolor{grey!10}{3.01}}&{\cellcolor{grey!10}{6.40}}&{221/0}&{217/0}&{204/2}&{169/13}\\
    
    {112}& {1.41e-7}&{7.25}&{8.55}&{5.24}&{4.91}&{\cellcolor{grey!10}{4.01}}&{\cellcolor{grey!10}{7.51}}&{222/0}&{217/0}&{202/2}&{166/16}\\
\hline
\end{tabular}
\caption{Evaluations of the $E5$ query encoder tuned with a frozen embedding block, margin 0.1 and 14000 samples per epoch. Square root scaling rule of learning rate with batch size. Values that did not pass the two-tailed test are shown in gray.}
\label{tab:Dependency_E5_batch_sqrt}
\end{table*}

The learning rate is usually set with consideration to the batch size; it can be proportional to the batch size (linear scaling rule), or proportional to square root of the batch size (square root scaling rule)~\cite{krizhevsky2014weirdtrickparallelizingconvolutional, goyal2018accuratelargeminibatchsgd, hoffer2018trainlongergeneralizebetter}. We show the evaluation results for these scaling rules in Tables~\ref{tab:Dependency_E5_batch_linear} and ~\ref{tab:Dependency_E5_batch_sqrt}. While there is no essential wins in scaling batch size and learning rate up or down, the square root rule seems more reasonable in keeping the evaluation results approximately the same while increasing the batch size.

Regardless of the overall behavior of scaling the batch size and learning rate together, we have to verify that our default batch size 14 is a good fit for our default learning rate $5e$-$8$. For this reason, a simple change of batch size, without altering learning rate, is considered in Appendix~\ref{app:E5_batch}; the tables~\ref{tab:Dependency_E5_batch_constNSamples} and \ref{tab:Dependency_E5_batch} show that our `default' batch size is reasonable. The corresponding data for $L12$ are in Appendix~\ref{app:L12_batch}.

\subsubsection{Encoder E5 and the batch size}\label{app:E5_batch}

\begin{table*}[th!]
\centering
\begin{tabular}{@{}c|rr|rr|rr|ll|ll@{}}
\toprule
{batch} & \multicolumn{2}{c|}{msmarco} & \multicolumn{2}{c|}{arxiv-first} & \multicolumn{2}{c|}{arxiv-title} & \multicolumn{2}{c|}{xnli ent-neutr} & \multicolumn{2}{c}{xnli ent-contr}\\
{size}&{c\%}&{d\%}&{c\%}&{d\%}&{c\%}&{d\%}&{c+/-}&{d+/-}&{c+/-}&{d+/-}\\
\hline
    {7}&{6.85}&{7.57}&{4.55}&{4.78}&{\cellcolor{grey!10}{3.01}}&{\cellcolor{grey!10}{6.62}}&{221/0}&{217/0}&{198/2}&{163/13}\\
    
    {14}&{7.30}&{8.82}&{4.90}&{5.39}&{\cellcolor{grey!10}{3.51}}&{\cellcolor{grey!10}{7.73}}&{222/0}&{217/1}&{201/2}&{158/20}\\
    
    {28}&{7.45}&{9.47}&{5.14}&{5.46}&{\cellcolor{grey!10}{4.01}}&{\cellcolor{grey!10}{8.39}}&{222/0}&{219/1}&{196/6}&{145/27}\\
    
    {56}&{6.51}&{7.33}&{\cellcolor{grey!10}{4.16}}&{4.48}&{\cellcolor{grey!10}{2.51}}&{\cellcolor{grey!10}{5.74}}&{221/0}&{217/0}&{202/1}&{168/6}\\
    
    {112}&{4.63}&{4.78}&{\cellcolor{grey!10}{2.55}}&{\cellcolor{grey!10}{2.50}}&{\cellcolor{grey!10}{2.51}}&{\cellcolor{grey!10}{2.21}}&{212/0}&{203/0}&{196/0}&{155/1}\\
\hline
\end{tabular}
\caption{Evaluations of the $E5$ query encoder tuned with a frozen embedding block, learning rate 5e-8, margin 0.1 and different batch sizes (first column); 14000 samples per epoch. Values that did not pass the two-tailed test are shown in gray.}
\label{tab:Dependency_E5_batch_constNSamples}
\end{table*}

\begin{table*}[th!]
\centering
\begin{tabular}{@{}c|rr|rr|rr|ll|ll@{}}
\toprule
{batch} & \multicolumn{2}{c|}{msmarco} & \multicolumn{2}{c|}{arxiv-first} & \multicolumn{2}{c|}{arxiv-title} & \multicolumn{2}{c|}{xnli ent-neutr} & \multicolumn{2}{c}{xnli ent-contr}\\
{size}&{c\%}&{d\%}&{c\%}&{d\%}&{c\%}&{d\%}&{c+/-}&{d+/-}&{c+/-}&{d+/-}\\
\hline
    {7}&{6.76}&{7.99}&{4.38}&{5.06}&{\cellcolor{grey!10}{3.26}}&{\cellcolor{grey!10}{6.18}}&{221/0}&{216/0}&{177/7}&{119/31}\\
    
    {14}&{7.30}&{8.82}&{4.90}&{5.39}&{\cellcolor{grey!10}{3.51}}&{\cellcolor{grey!10}{7.73}}&{222/0}&{217/1}&{201/2}&{158/20}\\
    
    {28}&{8.50}&{10.18}&{5.71}&{6.45}&{\cellcolor{grey!10}{2.76}}&{\cellcolor{grey!10}{8.17}}&{222/0}&{218/3}&{197/9}&{147/36}\\
    
    {56}&{7.42}&{9.12}&{4.55}&{4.81}&{\cellcolor{grey!10}{3.76}}&{\cellcolor{grey!10}{8.17}}&{223/0}&{220/0}&{200/2}&{160/21}\\
    
    {112}&{9.50}&{11.84}&{\cellcolor{grey!10}{-0.82}}&{\cellcolor{grey!10}{-4.05}}&{-15.54}&{-14.13}&{175/35}&{146/65}&{91/117}&{32/175}\\
\hline
\end{tabular}
\caption{Evaluations of the $E5$ query encoder tuned with a frozen embedding block, learning rate 5e-8, margin 0.1 and different batch sizes (first column); 1000 batches per epoch. Values that did not pass the two-tailed test are shown in gray.}
\label{tab:Dependency_E5_batch}
\end{table*}

In Table~\ref{tab:Dependency_E5_batch_constNSamples} we show results for batch sizes 7, 14, 28, 56 and 112, while keeping the number of samples per epoch the same (14000). The row with batch 14 here coincides with the values for learning rate 5e-8 in Figure~\ref{fig:e5_tuned_by_lr}, and with the row for the frozen embedding block in Table~\ref{tab:improve_E5_freeze}.
The results for all batch sizes are similar. Tuning with the higher batch size of 112 is a bit `safer' for languages, not degrading any language pair when evaluated by cosine measure, and degrading only one language pair (for entailment vs. contradiction) when evaluated by distance measure. This comes at the price of lower gains on MSMARCO and ARXIV.

\begin{table*}[th!]
\centering
\begin{tabular}{@{}c|rr|rr|rr|ll|ll@{}}
\toprule
{batch} & \multicolumn{2}{c|}{msmarco} & \multicolumn{2}{c|}{arxiv-first} & \multicolumn{2}{c|}{arxiv-title} & \multicolumn{2}{c|}{xnli ent-neutr} & \multicolumn{2}{c}{xnli ent-contr}\\
{size}&{c\%}&{d\%}&{c\%}&{d\%}&{c\%}&{d\%}&{c+/-}&{d+/-}&{c+/-}&{d+/-}\\
\hline
    {7}&{8.74}&{9.89}&{\cellcolor{grey!10}{1.59}}&{-16.43}&{9.59}&{-9.76}&{195/18}&{33/106}&{205/16}&{18/168}\\
    {14}&{7.28}&{8.04}&{\cellcolor{grey!10}{2.28}}&{-9.23}&{12.4}&{\cellcolor{grey!10}{-1.03}}&{200/15}&{47/47}&{206/15}&{20/143}\\
    {28}&{5.02}&{5.49}&{\cellcolor{grey!10}{1.98}}&{-4.05}&{8.37}&{\cellcolor{grey!10}{1.27}}&{199/15}&{36/27}&{196/15}&{7/77}\\
    {56}&{5.05}&{5.35}&{\cellcolor{grey!10}{1.73}}&{-4.3}&{8.66}&{\cellcolor{grey!10}{0.63}}&{197/15}&{35/28}&{195/15}&{6/89}\\
    {112}&{4.68}&{5.01}&{\cellcolor{grey!10}{1.69}}&{-3.64}&{7.47}&{\cellcolor{grey!10}{0.77}}&{193/15}&{25/25}&{188/15}&{4/86}\\
\hline
\end{tabular}
\caption{Evaluations of the $L12$ query encoder tuned with a frozen embedding block, learning rate 5e-8, margin 0.1 and different batch sizes (first column); 14000 samples per epoch. Values that did not pass the two-tailed test are shown in gray.}
\label{tab:L12_batch_constNSamples}
\end{table*}

\begin{table*}[th!]
\centering
\begin{tabular}{@{}c|rr|rr|rr|ll|ll@{}}
\toprule
{batch} & \multicolumn{2}{c|}{msmarco} & \multicolumn{2}{c|}{arxiv-first} & \multicolumn{2}{c|}{arxiv-title} & \multicolumn{2}{c|}{xnli ent-neutr} & \multicolumn{2}{c}{xnli ent-contr}\\
{size}&{c\%}&{d\%}&{c\%}&{d\%}&{c\%}&{d\%}&{c+/-}&{d+/-}&{c+/-}&{d+/-}\\
\hline
    {7}&{6.47}&{8.48}&{\cellcolor{grey!10}{-1.79}}&{-24.53}&{\cellcolor{grey!10}{-1.1}}&{-25.33}&{12/198}&{3/212}&{147/52}&{9/208}\\
    {14}&{7.28}&{8.04}&{\cellcolor{grey!10}{2.28}}&{-9.23}&{12.4}&{\cellcolor{grey!10}{-1.03}}&{200/15}&{47/47}&{206/15}&{20/143}\\
    {28}&{9.54}&{10.78}&{\cellcolor{grey!10}{1.11}}&{-20.67}&{9.16}&{-13.06}&{172/29}&{18/175}&{201/16}&{16/185}\\
    {56}&{9.51}&{10.86}&{\cellcolor{grey!10}{1.04}}&{-21.01}&{8.37}&{-13.63}&{173/28}&{18/175}&{201/16}&{17/181}\\
    {112}&{9.44}&{10.88}&{\cellcolor{grey!10}{1.06}}&{-20.86}&{8.18}&{-13.77}&{174/28}&{21/174}&{200/17}&{15/188}\\
\hline
\end{tabular}
\caption{Evaluations of the $L12$ query encoder tuned with a frozen embedding block, learning rate 5e-8, margin 0.1 and different batch sizes (first column); 1000 batches per epoch. Values that did not pass the two-tailed test are shown in gray.}
\label{tab:L12_batch}
\end{table*}

Table~\ref{tab:Dependency_E5_batch} shows what happens if the number of batches per epoch (1000) is kept the same, rather than the number of samples. In this setting the larger batch size of 112 leads to a less frequent validation (by MSMARCO validation subset) at tuning and, effectively, to later and less reasonable stopping. This results in higher gains on MSMARCO test subset, but in far worse results on ARXIV and XNLI.

\subsubsection{Encoder L12 and the batch size}\label{app:L12_batch}

The dependency of tuning $L12$ using different batch size is shown in Table~\ref{tab:L12_batch_constNSamples} (number of samples per epoch is 14000) and in Table~\ref{tab:L12_batch} (number of batches per epoch is 1000). Observations are somewhat similar to $E5$ (Appendix~\ref{app:E5_batch}), except that generally $L12$ does not perform as well as $E5$ and a batch size of 7 turns out to be bad for $L12$.

\subsection{Weight decay}\label{app:wdecay}

\begin{table*}[th!]
\centering
\begin{tabular}{@{}c|rr|rr|rr|ll|ll@{}}
\toprule
{weight} & \multicolumn{2}{c|}{msmarco} & \multicolumn{2}{c|}{arxiv-first} & \multicolumn{2}{c|}{arxiv-title} & \multicolumn{2}{c|}{xnli ent-neutr} & \multicolumn{2}{c}{xnli ent-contr}\\
{decay}&{c\%}&{d\%}&{c\%}&{d\%}&{c\%}&{d\%}&{c+/-}&{d+/-}&{c+/-}&{d+/-}\\
\hline
    {100}&{2.77}&{\cellcolor{grey!10}{1.88}}&{\cellcolor{grey!10}{-2.47}}&{\cellcolor{grey!10}{-1.06}}&{\cellcolor{grey!10}{4.01}}&{\cellcolor{grey!10}{1.32}}&{84/104}&{80/99}&{90/96}&{71/92}\\
    
    {50}&{5.39}&{5.00}&{\cellcolor{grey!10}{0.49}}&{\cellcolor{grey!10}{2.12}}&{\cellcolor{grey!10}{3.26}}&{\cellcolor{grey!10}{2.43}}&{120/51}&{121/41}&{140/51}&{133/46}\\
    
    {10}&{7.08}&{8.36}&{\cellcolor{grey!10}{3.54}}&{5.11}&{\cellcolor{grey!10}{4.26}}&{\cellcolor{grey!10}{6.62}}&{222/0}&{217/0}&{201/3}&{160/18}\\
    
    {5}&{7.88}&{9.84}&{4.97}&{5.87}&{\cellcolor{grey!10}{3.01}}&{\cellcolor{grey!10}{7.95}}&{222/0}&{216/2}&{189/15}&{144/36}\\
    
    {1}&{7.26}&{8.70}&{4.72}&{5.11}&{\cellcolor{grey!10}{3.01}}&{\cellcolor{grey!10}{7.06}}&{222/0}&{218/0}&{202/2}&{164/16}\\
    
    {0.5}&{7.28}&{8.78}&{4.87}&{5.24}&{\cellcolor{grey!10}{3.01}}&{\cellcolor{grey!10}{7.06}}&{221/0}&{218/0}&{202/2}&{163/18}\\
    
    {0.1}&{7.32}&{8.56}&{4.70}&{5.16}&{\cellcolor{grey!10}{3.01}}&{\cellcolor{grey!10}{6.40}}&{221/0}&{217/0}&{204/2}&{169/13}\\
    
    {0.05}&{7.32}&{8.56}&{4.70}&{5.16}&{\cellcolor{grey!10}{3.01}}&{\cellcolor{grey!10}{6.40}}&{221/0}&{217/0}&{204/2}&{169/13}\\
\hline
\end{tabular}
\caption{Evaluations of the $E5$ query encoder tuned with a frozen embedding block, learning rate 1e-7, batch size 56, margin 0.1 and a range of weight decay (first column). Values that did not pass the two-tailed test are shown in gray.}
\label{tab:E5_wdecay_batch56}
\end{table*}

\begin{table*}[th!]
\centering
\begin{tabular}{@{}c|rr|rr|rr|ll|ll@{}}
\toprule
{weight} & \multicolumn{2}{c|}{msmarco} & \multicolumn{2}{c|}{arxiv-first} & \multicolumn{2}{c|}{arxiv-title} & \multicolumn{2}{c|}{xnli ent-neutr} & \multicolumn{2}{c}{xnli ent-contr}\\
{decay}&{c\%}&{d\%}&{c\%}&{d\%}&{c\%}&{d\%}&{c+/-}&{d+/-}&{c+/-}&{d+/-}\\
\hline
    {5}&{6.53}&{7.26}&{\cellcolor{grey!10}{3.81}}&{4.50}&{\cellcolor{grey!10}{3.51}}&{\cellcolor{grey!10}{5.96}}&{222/0}&{216/0}&{197/2}&{150/14}\\
    
    {1}&{7.26}&{8.73}&{4.77}&{5.39}&{\cellcolor{grey!10}{3.76}}&{\cellcolor{grey!10}{7.95}}&{222/0}&{219/0}&{201/2}&{160/20}\\
    
    {0.5}&{7.30}&{8.82}&{4.90}&{5.39}&{\cellcolor{grey!10}{3.51}}&{\cellcolor{grey!10}{7.73}}&{222/0}&{217/1}&{201/2}&{158/20}\\
    
    {0.1}&{7.30}&{8.82}&{4.90}&{5.39}&{\cellcolor{grey!10}{3.51}}&{\cellcolor{grey!10}{7.73}}&{222/0}&{217/1}&{201/2}&{158/20}\\
\hline
\end{tabular}
\caption{Evaluations of the $E5$ query encoder tuned with a frozen embedding block, learning rate 5e-8, batch size 14, margin 0.1 and a range of weight decay (first column). Values that did not pass the two-tailed test are shown in gray.}
\label{tab:E5_wdecay_batch14}
\end{table*}

A weight decay may restrict increase of model weights, but it does not improve the evaluation results. We show some representative results in Tables~\ref{tab:E5_wdecay_batch56} and~\ref{tab:E5_wdecay_batch14}. While restricting gains on the tuning goal, weight decay does not help to preserve the other qualities: the results on XNLI and ARXIV are no better than without weight decay. If there is any recipe for further improving the gains both on the tuning goal and on the related qualities, it has to be a less crude interference into the tuning.

Since weight decay may be more effective at higher learning rates, the parameters for Table~\ref{tab:E5_wdecay_batch56} are chosen at higher rate and batch size, compared to our 'default' choice, which is used in Table~\ref{tab:E5_wdecay_batch14}. The learning rates and batch sizes of these tables relate by square root scaling rule (see Section~\ref{ssec:scaling_rule}).

\subsection{Candidate layers for freezing}\label{app:freeze_suspects}
In Section~\ref{ssec:extend_adiabatic_range} we showed how the adiabatic tuning range gets extended when the layer \textit{output.dense.weight} is frozen (in all blocks). The reason for suspecting that this layer is the most responsible for breaking out of the original `minimum' region, is that its maximal weight becomes the highest among all the layers as the learning rate gets closer to the end of the adiabatic range: see Table~\ref{tab:layers_suspect}. The maximal relative change of the weights is also achieved by the layer \textit{output.dense.weight}: see Table~\ref{tab:layers_suspect_2}. 

It is a crude adjustment, and freezing this layer in all blocks is probably overkill, but this did help us in extending the adiabatic range (Section~\ref{ssec:extend_adiabatic_range}).

\begin{table}[th!]
\centering
\begin{tabular}{@{}rl@{}}
\hline
{rate}&layer\\
\hline
{\multirow{2}{*}{1e-8}}&{1.intermediate.dense.bias}\\
{}&{3.intermediate.dense.bias}\\
\hline
{\multirow{2}{*}{2e-8}}&{5.intermediate.dense.weight}\\
{}&{3.attention.output.LayerNorm.weight}\\
\hline
{\multirow{2}{*}{3e-8}}&{5.intermediate.dense.weight}\\
{}&{1.attention.output.LayerNorm.weight}\\
\hline
{\multirow{2}{*}{4e-8}}&{5.intermediate.dense.weight}\\
{}&{3.attention.output.LayerNorm.weight}\\
\hline
{\multirow{2}{*}{5e-8}}&{3.output.dense.weight}\\
{}&{2.output.dense.weight}\\
\hline
{\multirow{2}{*}{6e-8}}&{3.output.dense.weight}\\
{}&{2.output.dense.weight}\\
\hline
{\multirow{2}{*}{7e-8}}&{3.output.dense.weight}\\
{}&{2.output.dense.weight}\\
\hline
{\multirow{2}{*}{8e-8}}&{1.output.dense.weight}\\
{}&{3.output.dense.weight}\\
\hline
{\multirow{2}{*}{9e-8}}&{1.output.dense.weight}\\
{}&{4.output.dense.weight}\\
\hline
{\multirow{2}{*}{1e-7}}&{1.output.dense.weight}\\
{}&{5.output.dense.weight}\\
\hline

\end{tabular} 
\caption{The 'most changed' two layers at each learning rate.
The 'change' is defined as the maximal weight of the layer \textit{if} it was changed by the tuning. The prefix 'encoder.layer' is removed from the layer names here. 
}
\label{tab:layers_suspect}
\end{table}

\begin{table}[th!]
\centering
\begin{tabular}{@{}rl@{}}
\hline
{rate}&layer\\
\hline
{\multirow{2}{*}{1e-8}}&{5.attention.output.dense.bias}\\
{}&{11.output.dense.bias}\\
\hline
{\multirow{2}{*}{2e-8}}&{5.attention.output.dense.bias}\\
{}&{11.output.dense.bias}\\
\hline
{\multirow{2}{*}{3e-8}}&{5.attention.output.dense.bias}\\
{}&{11.output.dense.bias}\\
\hline
{\multirow{2}{*}{4e-8}}&{5.attention.output.dense.bias}\\
{}&{11.output.dense.bias}\\
\hline
{\multirow{2}{*}{5e-8}}&{11.output.dense.weight}\\
{}&{11.output.dense.bias}\\
\hline
{\multirow{2}{*}{6e-8}}&{11.output.dense.weight}\\
{}&{11.output.dense.bias}\\
\hline
{\multirow{2}{*}{7e-8}}&{11.output.dense.weight}\\
{}&{11.output.dense.bias}\\
\hline
{\multirow{2}{*}{8e-8}}&{11.output.dense.weight}\\
{}&{11.output.dense.bias}\\
\hline
{\multirow{2}{*}{9e-8}}&{11.output.dense.weight}\\
{}&{11.output.dense.bias}\\
\hline
{\multirow{2}{*}{1e-7}}&{11.output.dense.weight}\\
{}&{11.output.dense.bias}\\
\hline
\end{tabular} 
\caption{The 'most changed' two layers at each learning rate.
The 'change' is defined as $(W_t - W_o)/(W_t + W_o)$, where $W_t$ is the maximal weight of the layer in the tuned query encoder, and $W_o$ is the maximal weight of the layer in the original (untuned) encoder. The prefix 'encoder.layer' is removed from the layer names here.
}
\label{tab:layers_suspect_2}
\end{table}

\subsection{Margin of triple loss}\label{app:E5_margin}
\begin{figure}[h!]
    \centering
    \includegraphics[width=\linewidth]{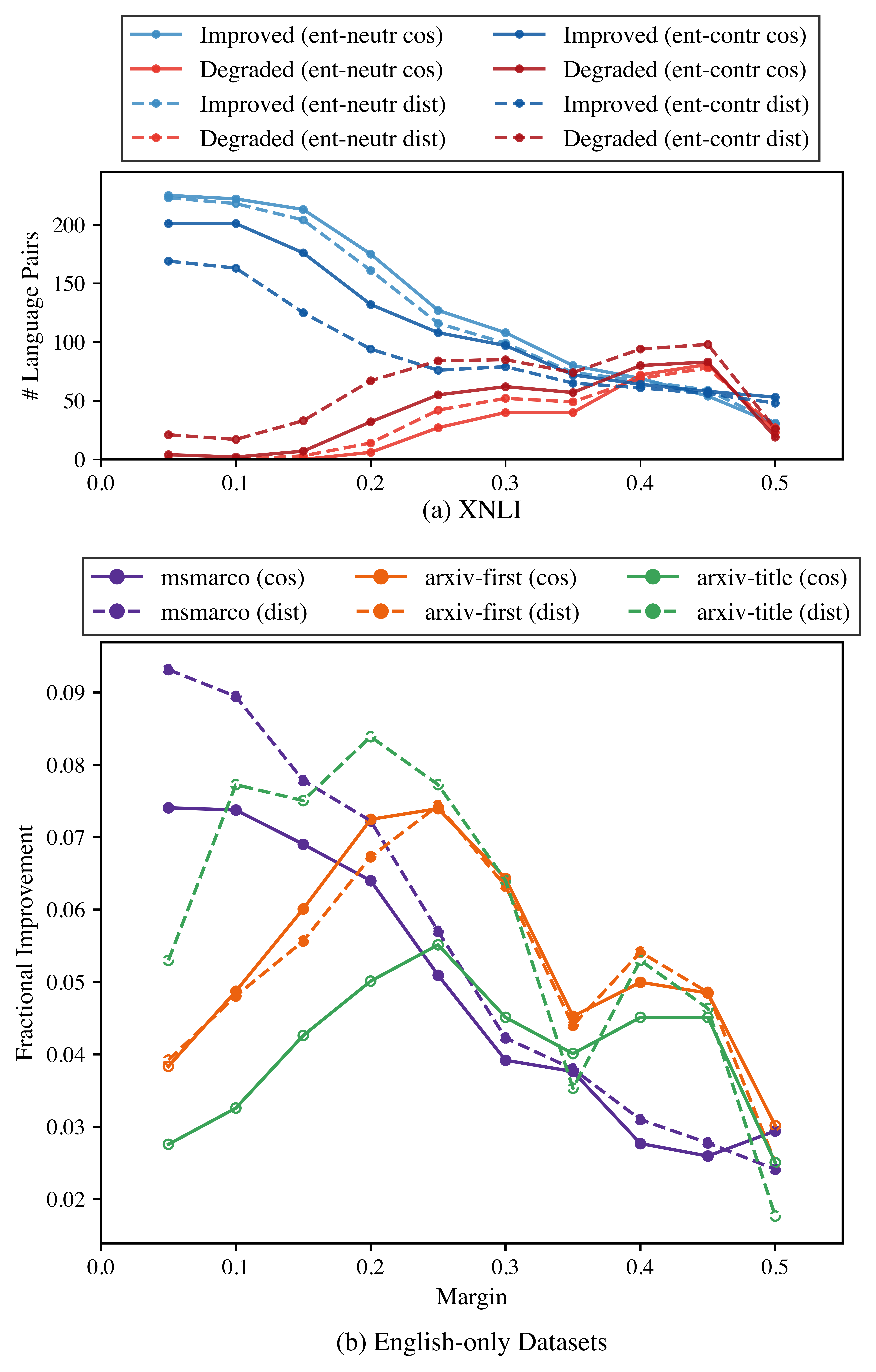}
    \caption{Evaluations of the $E5$ query encoder tuned with a frozen embedding block, learning rate 5e-8, batch size 14 using different triplet loss margins on (a) XNLI and (b) the English-only datasets (MSMARCO and ARXIV). Values that did not pass the two-tailed test are shown with open markers.}
    \label{fig:E5_margin}
\end{figure}

When using the triplet loss for contrastive learning, the margin is an important parameter that can significantly affect model training. In Figure~\ref{fig:E5_margin} we show the dependency of the evaluation results on the margin during its tuning. We consider our default tuning parameters (Section~\ref{ssec:Tuning}), but change the margin. The results are not unexpected: a margin up to 0.15 is reasonable, and at higher margins the disturbance on cross-lingual, and, eventually, on English data evaluation becomes too strong.

\begin{figure}[h!]
    \centering
    \includegraphics[width=\linewidth]{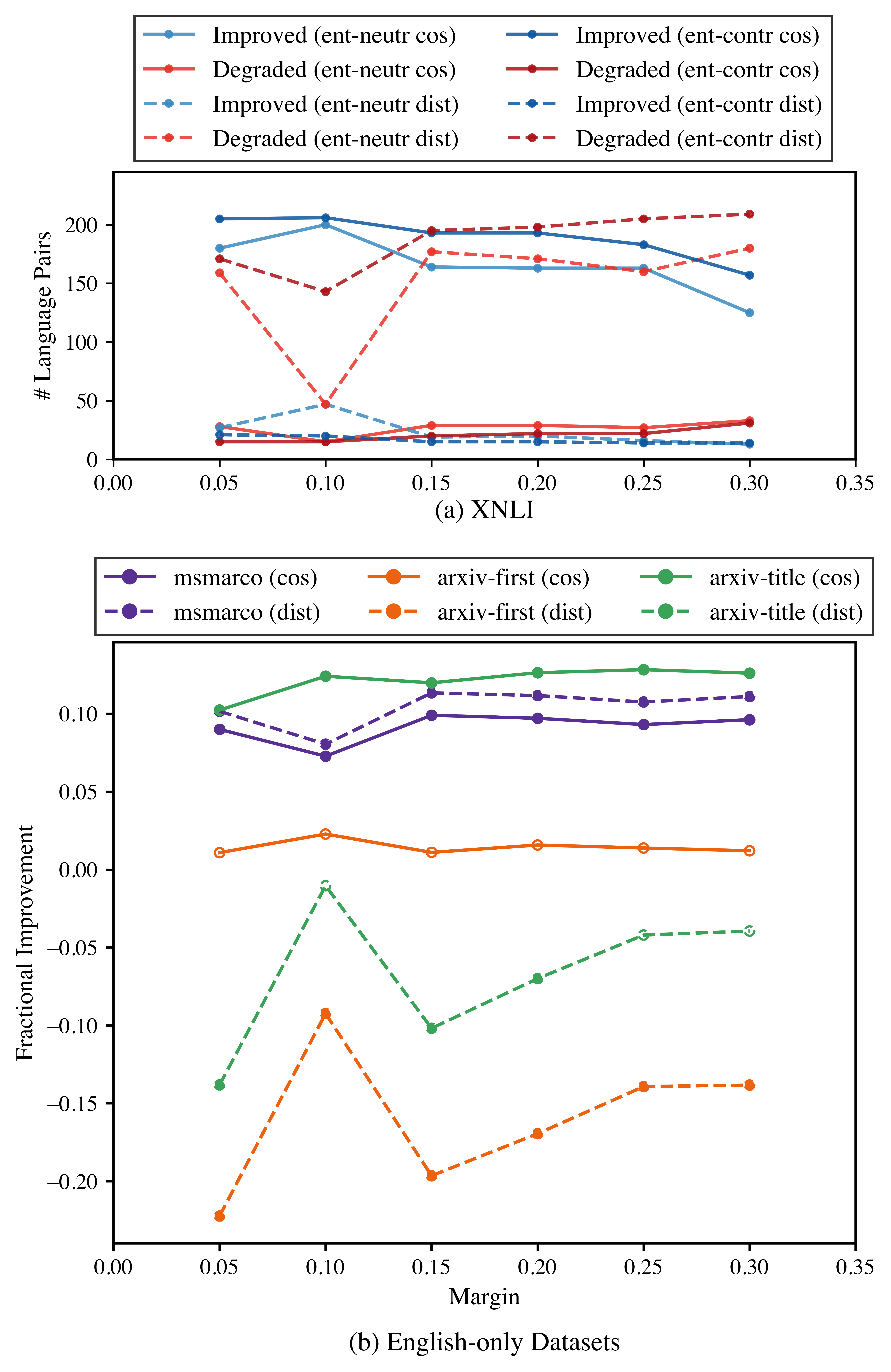}
    \caption{Evaluations of the $L12$ query encoder tuned with a frozen embedding block, learning rate 5e-8, batch size 14 using different triplet loss margins on (a) XNLI and (b) the English-only datasets (MSMARCO and ARXIV). Values that did not pass the two-tailed test are shown with open markers.}
    \label{fig:L12_margin}
\end{figure}

The corresponding data for $L12$ are given in Figure~\ref{fig:L12_margin}. It shows that a margin of 0.1 works best for $L12$. The results for margin 0.1 are distinctly better. Altogether, $L12$ appears to be more sensitive (compared to $E5$) to the tuning parameters if the goal is to preserve performance on multilingual XNLI data and on out-of-domain ARXIV data. Arguably, the margin value of approximately 0.1 is the best both for $L12$ and $E5$.

\subsection{Stopping criterion}\label{app:E5_stopping}

In Table~\ref{tab:Dependency_E5_stopping} we show how the improvement depends on the stopping criterion. The stoppings after 5 or 10 non-improvement epochs give similar results. Stopping after 15 non-improvement epochs continues the trend of increased gain on English data, but with a deterioration on a few language pairs.

\begin{table*}[th!]
\centering
\begin{tabular}{@{}c|rr|rr|rr|ll|ll@{}}
\toprule
{idle epochs} & \multicolumn{2}{c|}{msmarco} & \multicolumn{2}{c|}{arxiv-first} & \multicolumn{2}{c|}{arxiv-title} & \multicolumn{2}{c|}{xnli ent-neutr} & \multicolumn{2}{c}{xnli ent-contr}\\
{to stop}&{c\%}&{d\%}&{c\%}&{d\%}&{c\%}&{d\%}&{c+/-}&{d+/-}&{c+/-}&{d+/-}\\
\hline
    {5}&{6.50}&{7.60}&{\cellcolor{grey!10}{4.18}}&{4.40}&{\cellcolor{grey!10}{2.26}}&{\cellcolor{grey!10}{6.62}}&{222/0}&{217/0}&{208/1}&{174/5}\\
    
    {10}&{7.38}&{8.95}&{4.87}&{4.81}&{\cellcolor{grey!10}{3.26}}&{\cellcolor{grey!10}{7.73}}&{222/0}&{218/0}&{201/2}&{163/17}\\
    
    {15}&{8.93}&{10.98}&{5.81}&{6.10}&{\cellcolor{grey!10}{2.51}}&{\cellcolor{grey!10}{7.73}}&{222/0}&{217/3}&{191/17}&{140/51}\\
\hline
\end{tabular}
\caption{Evaluations of the $E5$ query encoder tuned with a frozen embedding block, learning rate 5e-8, batch size 14 and triplet loss margin 0.1, stopped after different number of idle epochs (first column). The epoch is idle if no improvement is made. Values that did not pass the two-tailed test are shown in gray.}
\label{tab:Dependency_E5_stopping}
\end{table*}

\subsection{Execution time}\label{app:time}
There is no essential difference between the execution times for $E5$ and $L12$.
The tuning time depends on how soon stopping happened. At the settings of interest (Section~\ref{ssec:Tuning},~\ref{ssec:freezing},~\ref{ssec:adiabatic tuning}), the tuning on an A100 GPU takes about one hour. For example, tuning 10 times at the default settings (Section~\ref{ssec:Tuning}, Appendix~\ref{app:tuning}) for rates between 1e-8 and 1e-7 takes 9 hours. At higher rates, stopping occurs earlier; tuning 10 times for rates between 1.1e-7 to 2e-7 takes less than 5 hours. Table~\ref{tab:improve_E5_freeze} (with freezing different parts of the encoder) was obtained in 6 hours.

Evaluation of an encoder on all datasets we used (MSMARCO, ARXIV-first, ARXIV-title and XNLI) takes about 1.2-1.3 hours.

\subsection{Effects of learning rate scheduler and weight decay}\label{app:scheduling}

\begin{table*}[th!]
    \centering
    \begin{tabular}{@{}clc|rr|rr|rr|ll|ll@{}}
        \toprule
        {} & {} & {} & \multicolumn{2}{c|}{msmarco} & \multicolumn{2}{c|}{arxiv-first} & \multicolumn{2}{c|}{arxiv-title} & \multicolumn{2}{c|}{xnli ent-neutr} & \multicolumn{2}{c}{xnli ent-contr}\\
        {B} & {Sch} & {D} & {c\%} & {d\%} & {c\%} & {d\%} & {c\%} & {d\%} & {c+/-} & {d+/-} & {c+/-} & {d+/-} \\[0.125cm]
        \hline\hline
        \parbox[t]{2mm}{\multirow{2}{*}{\rotatebox[origin=c]{90}{100}}}
        & $Q$ & - & \chl{teal}{65}{1.83} & \chl{teal}{65}{2.80} & \chl{teal}{65}{0.99} & \chl{teal}{65}{1.23} & -0.26 & \chl{teal}{40}{0.96} & \chl{teal}{40}{200/1} & \chl{teal}{40}{172/3} & \chl{red}{65}{69/71} & \chl{teal}{20}{43/110} \\[0.125cm]
        & $E_{0.98}$ & $10^{-6}$ & -0.41 & -1.07 & -0.29 & -0.83 & -0.26 & -0.96 & \chl{red}{20}{0/63} & \chl{red}{20}{0/39} & \chl{red}{20}{0/20} & 1/8 \\
        \hline
        \parbox[t]{2mm}{\multirow{7}{*}{\rotatebox[origin=c]{90}{64}}}
        & - & - & \chl{teal}{40}{1.58} & \chl{teal}{40}{2.28} & \chl{teal}{40}{0.78} & \chl{teal}{40}{1.07} & \chl{teal}{65}{0.78} & \chl{teal}{65}{1.20} & \chl{teal}{20}{178/1} & \chl{teal}{20}{156/3} & \chl{teal}{40}{72/50} & \chl{teal}{40}{48/87} \\[0.125cm]
        & $Q$ & - & 0.05 & 0.19 & -0.34 & -0.59 & -0.26 & 0.00 & \chl{teal}{65}{0/0} & \chl{teal}{65}{0/0} & \chl{teal}{65}{0/0} & \chl{red}{20}{0/1} \\[0.125cm]
        & $E_{0.98}$ & $10^{-6}$ & 0.13 & 0.17 & -0.23 & -0.32 & -0.26 & 0.00 & \chl{teal}{65}{0/0} & \chl{teal}{65}{0/0} & \chl{teal}{65}{0/0} & \chl{red}{20}{0/1} \\[0.125cm]
        & $E_{0.95}$ & $10^{-6}$ & \chl{red}{65}{-0.75} & \chl{red}{40}{-1.47} & -0.62 & -0.78 & -0.52 & -0.48 & \chl{red}{20}{0/67} & \chl{red}{20}{0/48} & \chl{red}{20}{0/66} & \chl{red}{40}{1/42} \\[0.125cm]
        & $E_{0.95}$ & $10^{-5}$ & \chl{red}{65}{-0.75} & \chl{red}{40}{-1.47} & -0.62 & -0.78 & -0.52 & -0.48 & \chl{red}{20}{0/67} & \chl{red}{20}{0/48} & \chl{red}{20}{0/66} & \chl{red}{40}{1/42} \\[0.125cm]
        & - & $10^{-4}$ & \chl{teal}{40}{1.58} & \chl{teal}{40}{2.28} & \chl{teal}{40}{0.78} & \chl{teal}{40}{1.07} & \chl{teal}{65}{0.78} & \chl{teal}{65}{1.20} & \chl{teal}{20}{178/1} & \chl{teal}{20}{156/3} & \chl{teal}{40}{72/50} & \chl{teal}{40}{48/87} \\[0.125cm]
        & $L$ & $10^{-4}$ & \chl{teal}{20}{0.30} & \chl{teal}{20}{0.55} & -0.31 & \chl{teal}{20}{-0.11} & \chl{red}{65}{-0.78} & -0.48 & \chl{teal}{65}{0/0} & \chl{teal}{65}{0/0} & \chl{red}{20}{0/4} & \chl{red}{20}{0/9} \\
        \hline
        \parbox[t]{2mm}{\multirow{3}{*}{\rotatebox[origin=c]{90}{32}}}
        & $Q$ & $10^{-4}$ & 0.12 & -0.17 & -0.29 & -0.43 & \chl{teal}{20}{0.26} & 0.00 & \chl{teal}{65}{0/0} & \chl{teal}{65}{0/0} & \chl{teal}{65}{0/0} & \chl{teal}{65}{0/0} \\[0.125cm]
        & $L$ & $10^{-4}$ & -0.05 & -0.21 & -0.18 & -0.19 & \chl{teal}{65}{0.78} & 0.00 & \chl{teal}{65}{0/0} & \chl{teal}{65}{0/0} & \chl{teal}{65}{0/0} & \chl{teal}{65}{0/0} \\[0.125cm]
        & $E_{0.98}$ & $10^{-4}$ & 0.27 & 0.46 & -0.21 & -0.16 & \chl{teal}{40}{0.52} & \chl{teal}{20}{0.24} & \chl{teal}{65}{26/0} & \chl{teal}{65}{39/0} & \chl{teal}{65}{0/0} & \chl{teal}{65}{0/0} \\
        \hline
        \parbox[t]{2mm}{\multirow{3}{*}{\rotatebox[origin=c]{90}{16}}}
        & $L$ & - & 0.00 & -0.15 & \chl{teal}{20}{-0.10} & -0.56 & \chl{teal}{65}{0.78} & \chl{teal}{20}{0.24} & \chl{teal}{65}{0/0} & \chl{teal}{65}{0/0} & \chl{teal}{65}{4/0} & \chl{teal}{65}{10/0} \\[0.125cm]
        & $E_{0.95}$ & - & -0.70 & -1.21 & \chl{red}{65}{-0.65} & -0.78 & \chl{teal}{20}{0.26} & -0.96 & \chl{red}{20}{0/53} & \chl{red}{20}{0/32} & \chl{teal}{20}{7/6} & \chl{teal}{65}{20/0} \\[0.125cm]
        & $E_{0.95}$ & $10^{-4}$ & -0.70 & -1.21 & \chl{red}{65}{-0.65} & -0.78 & \chl{teal}{20}{0.26} & -0.96 & \chl{red}{20}{0/53} & \chl{red}{20}{0/32} & \chl{teal}{20}{7/6} & \chl{teal}{65}{20/0} \\
        \hline
        \parbox[t]{2mm}{\multirow{5}{*}{\rotatebox[origin=c]{90}{8}}}
        & $Q$ & $10^{-6}$ & \chl{red}{40}{-0.81} & -1.39 & -0.57 & \chl{red}{65}{-1.15} & \chl{red}{40}{-1.30} & \chl{red}{65}{-2.39} & \chl{red}{20}{0/167} & \chl{red}{20}{0/123} & \chl{red}{20}{0/121} & \chl{red}{65}{2/81} \\
        & $E_{0.95}$ & $10^{-6}$ & \chl{red}{20}{-2.98} & \chl{red}{20}{-4.31} & \chl{red}{20}{-2.34} & \chl{red}{20}{-2.73} & \chl{red}{20}{-2.60} & \chl{red}{20}{-6.22} & \chl{red}{20}{0/220} & \chl{red}{20}{0/208} & \chl{red}{40}{2/187} & 15/128 \\[0.125cm]
        & - & $10^{-5}$ & \chl{red}{40}{-0.81} & \chl{red}{65}{-1.43} & \chl{red}{40}{-0.78} & \chl{red}{40}{-1.18} & \chl{red}{40}{-1.30} & \chl{red}{40}{-2.87} & \chl{red}{20}{0/165} & \chl{red}{20}{0/126} & \chl{red}{20}{0/115} & 3/68 \\[0.125cm]
        & $Q$ & $10^{-4}$ & \chl{red}{40}{-0.81} & -1.39 & -0.57 & \chl{red}{65}{-1.15} & \chl{red}{40}{-1.30} & \chl{red}{65}{-2.39} & \chl{red}{20}{0/167} & \chl{red}{20}{0/123} & \chl{red}{20}{0/121} & \chl{red}{65}{2/81} \\[0.125cm]
        & $E_{0.95}$ & $10^{-4}$ & \chl{red}{20}{-2.98} & \chl{red}{20}{-4.31} & \chl{red}{20}{-2.34} & \chl{red}{20}{-2.73} & \chl{red}{20}{-2.60} & \chl{red}{20}{-6.22} & \chl{red}{20}{0/220} & \chl{red}{20}{0/208} & \chl{red}{40}{2/187} & 15/128 \\
        \bottomrule
    \end{tabular}
    \caption{Percentage improvement over the fine-tuned E5 model with a frozen embedding block and tuned using a batch size of $14$, learning rate $5e$-$8$ and a margin of $0.1$.  The blue colors indicate the top improvements whereas the red colors indicate the worse degradation. Three parameters are randomly varied: the batch size (denoted as ``B''), the learning rate scheduler (denoted as ``Sch'') and the weight decay (denoted as ``D'').  The learning rate schedulers are defined in Table~\ref{tab:scheduler} with an initial learning rate of $5e$-$8$.  c\% and d\% refer to measuring the similarity of the text pairs using either the cosine similarity or the euclidean distance, respectively.  For XNLI, $(+)$ indicates the number of language pairs that were improved while $(-)$ indicates those that have worsened out of a total of 225 language pairs.  Note that only the statistically significant (determined by a Z-test) language pairs are retained and hence not all the improved/worsened counts sum to 225.  Additionally, (ent-neutr) refers to entailment-entailment similarities compared with entailment-neutral similarities whereas (ent-contr) refers to comparisons against entailment-contradiction similarities.} 
    \label{tab:hsweep2}
\end{table*}

\begin{table*}[th!]
    \centering
    \begin{tabular}{@{}clc|rr|rr|rr|ll|ll@{}}
        \toprule
        {} & {} & {} & \multicolumn{2}{c|}{msmarco} & \multicolumn{2}{c|}{arxiv-first} & \multicolumn{2}{c|}{arxiv-title} & \multicolumn{2}{c|}{xnli ent-neutr} & \multicolumn{2}{c}{xnli ent-contr}\\
        {B} & {Sch} & {D} & {c\%} & {d\%} & {c\%} & {d\%} & {c\%} & {d\%} & {c+/-} & {d+/-} & {c+/-} & {d+/-} \\[0.125cm]
        \hline\hline
        \parbox[t]{2mm}{\multirow{2}{*}{\rotatebox[origin=c]{90}{100}}}
        & $Q$ & - & -1.12 & -1.17 & -0.82 & \chl{teal}{20}{-0.14} & \chl{red}{40}{-0.52} & -1.71 & 0/138 & 3/117 & 15/65 & 61/40 \\[0.125cm]
        & $E_{0.98}$ & $10^{-6}$ & \chl{teal}{40}{-0.31} & \chl{teal}{40}{-0.18} & \chl{teal}{65}{-0.39} & \chl{teal}{65}{-0.05} & \chl{red}{40}{-0.52} & \chl{teal}{40}{-1.22} & \chl{teal}{65}{0/0} & \chl{teal}{65}{0/0} & \chl{teal}{40}{0/11} & 0/7 \\
        \hline
        \parbox[t]{2mm}{\multirow{7}{*}{\rotatebox[origin=c]{90}{64}}}
        & - & - & -1.20 & -1.05 & -0.79 & \chl{teal}{40}{-0.11} & \chl{teal}{20}{-0.26} & \chl{teal}{65}{-0.73} & \chl{teal}{40}{0/57} & \chl{teal}{40}{5/34} & 1/80 & \chl{red}{65}{9/47} \\[0.125cm]
        & $Q$ & - & \chl{teal}{20}{-0.72} & -0.89 & \chl{teal}{20}{-0.66} & \chl{teal}{20}{-0.14} & \chl{teal}{20}{-0.26} & -1.71 & 0/65 & \chl{teal}{20}{4/43} & 1/62 & 21/33 \\[0.125cm]
        & $E_{0.98}$ & $10^{-6}$ & -0.96 & -1.13 & \chl{teal}{20}{-0.66} & -0.49 & \chl{red}{40}{-0.52} & \chl{teal}{40}{-1.22} & 0/80 & 6/56 & 3/65 & 29/35 \\[0.125cm]
        & $E_{0.95}$ & $10^{-6}$ & -1.78 & -2.36 & -0.97 & -1.35 & \chl{teal}{65}{0.78} & \chl{teal}{20}{-1.46} & 1/160 & 9/131 & \chl{red}{20}{22/103} & 73/68 \\[0.125cm]
        & $E_{0.95}$ & $10^{-5}$ & -1.78 & -2.36 & -0.97 & -1.35 & \chl{teal}{65}{0.78} & \chl{teal}{20}{-1.46} & 1/160 & 9/131 & \chl{red}{20}{22/103} & 73/68 \\[0.125cm]
        & - & $10^{-4}$ & -1.20 & -1.05 & -0.79 & \chl{teal}{40}{-0.11} & \chl{teal}{20}{-0.26} & \chl{teal}{65}{-0.73} & \chl{teal}{40}{0/57} & \chl{teal}{40}{5/34} & 1/80 & \chl{red}{65}{9/47} \\[0.125cm]
        & $L$ & $10^{-4}$ & -0.76 & \chl{teal}{20}{-0.80} & \chl{teal}{40}{-0.58} & -0.16 & \chl{teal}{20}{-0.26} & \chl{teal}{20}{-1.46} & \chl{teal}{20}{0/61} & \chl{teal}{40}{3/32} & 0/58 & 14/32 \\
        \hline
        \parbox[t]{2mm}{\multirow{3}{*}{\rotatebox[origin=c]{90}{32}}}
        & $Q$ & $10^{-4}$ & \chl{red}{20}{-2.12} & -3.35 & \chl{red}{40}{-1.47} & \chl{red}{20}{-1.60} & \chl{teal}{40}{0.00} & \chl{red}{20}{-2.68} & 2/190 & 8/162 & 41/107 & 93/69 \\[0.125cm]
        & $L$ & $10^{-4}$ & -2.05 & \chl{red}{20}{-3.36} & \chl{red}{20}{-1.45} & -1.33 & \chl{teal}{40}{0.00} & \chl{red}{20}{-2.68} & 2/189 & 8/158 & 41/102 & \chl{teal}{20}{94/68} \\[0.125cm]
        & $E_{0.98}$ & $10^{-4}$ & \chl{red}{20}{-2.12} & \chl{red}{40}{-3.54} & -1.42 & \chl{red}{65}{-1.84} & \chl{teal}{40}{0.00} & \chl{red}{20}{-2.68} & 2/192 & 8/163 & 41/110 & 93/71 \\
        \hline
        \parbox[t]{2mm}{\multirow{3}{*}{\rotatebox[origin=c]{90}{16}}}
        & $L$ & - & \chl{teal}{65}{-0.02} & \chl{teal}{65}{0.13} & \chl{teal}{40}{-0.58} & -0.65 & \chl{red}{65}{-1.04} & \chl{teal}{40}{-1.22} & \chl{teal}{65}{0/0} & \chl{teal}{65}{0/0} & \chl{teal}{65}{11/0} & \chl{teal}{40}{43/0} \\[0.125cm]
        & $E_{0.95}$ & - & \chl{red}{65}{-2.52} & -3.32 & -1.32 & -1.38 & \chl{teal}{20}{-0.26} & -2.20 & 3/186 & \chl{red}{20}{8/164} & \chl{teal}{20}{49/86} & \chl{teal}{65}{101/55} \\[0.125cm]
        & $E_{0.95}$ & $10^{-4}$ & \chl{red}{65}{-2.52} & -3.32 & -1.32 & -1.38 & \chl{teal}{20}{-0.26} & -2.20 & 3/186 & \chl{red}{20}{8/164} & \chl{teal}{20}{49/86} & \chl{teal}{65}{101/55} \\
        \hline
        \parbox[t]{2mm}{\multirow{5}{*}{\rotatebox[origin=c]{90}{8}}}
        & $Q$ & $10^{-6}$ & -2.05 & -3.29 & -1.11 & -1.38 & \chl{teal}{20}{-0.26} & \chl{red}{65}{-3.66} & \chl{red}{40}{0/212} & \chl{red}{65}{3/182} & \chl{red}{65}{21/140} & \chl{red}{40}{70/105} \\[0.125cm]
        & $E_{0.95}$ & $10^{-6}$ & \chl{red}{40}{-2.44} & \chl{red}{65}{-3.81} & \chl{red}{65}{-1.55} & \chl{red}{40}{-1.78} & \chl{red}{40}{-0.52} & \chl{red}{40}{-3.41} & \chl{red}{65}{0/213} & \chl{red}{40}{4/182} & \chl{red}{40}{24/135} & 76/97 \\[0.125cm]
        & - & $10^{-5}$ & -2.03 & -3.19 & -0.74 & -1.57 & \chl{teal}{20}{-0.26} & \chl{red}{40}{-3.41} & \chl{red}{20}{0/209} & \chl{red}{65}{3/182} & \chl{red}{65}{20/139} & \chl{red}{20}{71/103} \\[0.125cm]
        & $Q$ & $10^{-4}$ & -2.05 & -3.29 & -1.11 & -1.38 & \chl{teal}{20}{-0.26} & \chl{red}{65}{-3.66} & \chl{red}{40}{0/212} & \chl{red}{65}{3/182} & \chl{red}{65}{21/140} & \chl{red}{40}{70/105} \\[0.125cm]
        & $E_{0.95}$ & $10^{-4}$ & \chl{red}{40}{-2.44} & \chl{red}{65}{-3.81} & \chl{red}{65}{-1.55} & \chl{red}{40}{-1.78} & \chl{red}{40}{-0.52} & \chl{red}{40}{-3.41} & \chl{red}{65}{0/213} & \chl{red}{40}{4/182} & \chl{red}{40}{24/135} & 76/97 \\
        \bottomrule
    \end{tabular}
    \caption{Percentage improvement over the fine-tuned E5 model with a frozen embedding block and tuned using a batch size of $14$, learning rate $10^{-7}$ and a margin of $0.1$.  The blue colors indicate the top improvements whereas the red colors indicate the worse degradation. Three parameters are randomly varied: the batch size (denoted as ``B''), the learning rate scheduler (denoted as ``Sch'') and the weight decay (denoted as ``D'').  The learning rate schedulers are defined in Table~\ref{tab:scheduler} with an initial learning rate of $10^{-7}$.  c\% and d\% refer to measuring the similarity of the text pairs using either the cosine similarity or the euclidean distance, respectively.  For XNLI, $(+)$ indicates the number of language pairs that were improved while $(-)$ indicates those that have worsened out of a total of 225 language pairs.  Note that only the statistically significant (determined by a Z-test) language pairs are retained and hence not all the improved/worsened counts sum to 225.  Additionally, (ent-neutr) refers to entailment-entailment similarities compared with entailment-neutral similarities whereas (ent-contr) refers to comparisons against entailment-contradiction similarities.} 
    \label{tab:hsweep}
\end{table*}

\begin{table}[]
    \centering
    \begin{tabular}{l c}
        \toprule
        Scheduler & Definition \\
        \hline\hline
        $L$ & $\alpha(t) = \alpha_0 \left(1 - \frac{t}{T}\right)$ \\[0.125cm]
        $Q$ & $\alpha(t) = \alpha_0 \left(1 - \left(\frac{t}{T}\right)^2\right)$ \\[0.125cm]
        $E_{0.95}$ & $\alpha(t) = 0.95^t \alpha_0$ \\[0.125cm]
        $E_{0.98}$ & $\alpha(t) = 0.98^t \alpha_0$ \\
        \bottomrule
    \end{tabular}
    \caption{The definitions of the various learning rate schedulers used in Table~\ref{tab:hsweep} where $t$ is the current training step, $T$, the total number of training steps and $\alpha_0$, the initial learning rate.}
    \label{tab:scheduler}
\end{table}

Using the fine-tuned $E5$ model with the frozen embedding block, tuned using a batch size of 14, and a margin of 0.1, we randomly vary the batch size, learning rate scheduler and weight decay in order to assess their impact on the model's final performance.  In Table~\ref{tab:hsweep2} we present the change in performance across these different configurations for a learning rate of $5 \times 10^{-8}$, which is our `default' learning rate (Section~\ref{ssec:Tuning}). In Table~\ref{tab:hsweep} we do the same for a learning rate of $10^{-7}$. Table~\ref{tab:scheduler} lists the different schedulers we considered.  Values in blue indicate the top improvements whereas values in red indicate the worse degradation.

Across these parameters, on average, the batch size appears to have the most significant impact, generally leading to poorer performance as the batch size is decreased.  Within each batch size group, we see that using an exponential learning rate scheduler ($E_{0.95}$ or $E_{0.98}$) is generally worse than using any of the other schedulers or no scheduler at all.  A specific exception exists when using a batch size of 100 where the exponential scheduler outperforms the quadratic one when the learning rate is set to $10^{-7}$.  Across all the configurations considered, the most impact seems to be the one shown in the first row of Table~\ref{tab:hsweep2}, where we see good improvement over MSMARCO and ARXIV-first while simultaneously showing improvement over XNLI ent-neutr.

\subsection{Varying the optimizer and learning rate}\label{app:optim}

\begin{figure}[h!]
    \centering
    \includegraphics[width=\linewidth]{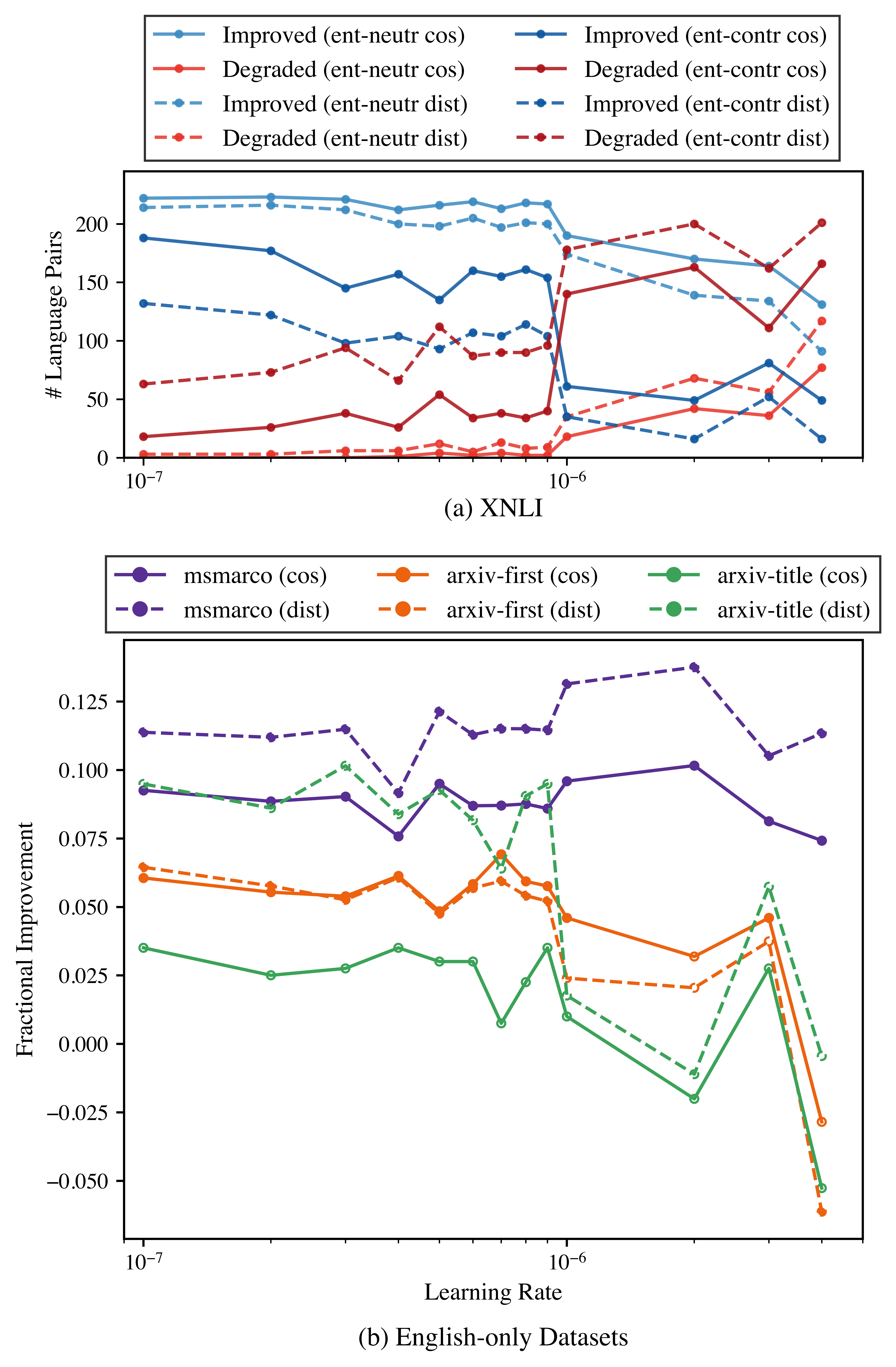}
    \caption{Evaluations on (a) XNLI and (b) the English-only datasets (MSMARCO and ARXIV) of the E5 query encoder tuned with a frozen embedding block, batch size 14, margin 0.1 using different learning rates.  Here we tune using SGD without momentum.  Values that did not pass the two-tailed test are shown with open markers.}
    \label{fig:sgd_lrs}
\end{figure}

Table~\ref{tab:optim} shows the effects of choosing a different optimizer with a small and large learning rate.  In addition to Adamax, we tried Adadelta and Stochastic Gradient Descent (SGD), both of which did not change the model weights in a significant enough way to affect the overall performance and hence, are not presented.  For higher learning rates, SGD without momentum did elicit a change as shown in Fig.~\ref{fig:sgd_lrs}, but the trend in performance is similar to what is presented in Fig.~\ref{fig:e5_tuned_by_lr} with higher resolution near the transition point between improved and degraded multilingual performance.  At around $9\times10^{-7}$, we see a sharp increase in the number of degraded language pairs while the model maintains constant improvement on MSMARCO.  With a high enough learning rate, it seems that the gradients are able to overcome a barrier in the loss landscape that confined the weights to a region in which multilingual characteristics were preserved.

\begin{table*}[t]
    \centering
    \begin{tabular}{@{}clc|rr|rr|rr|ll|ll@{}}
        \toprule
        {} & {} & {} & \multicolumn{2}{c|}{msmarco} & \multicolumn{2}{c|}{arxiv-first} & \multicolumn{2}{c|}{arxiv-title} & \multicolumn{2}{c|}{xnli ent-neutr} & \multicolumn{2}{c}{xnli ent-contr}\\
        {O} & {M} & {LR} & {c\%} & {d\%} & {c\%} & {d\%} & {c\%} & {d\%} & {c+/-} & {d+/-} & {c+/-} & {d+/-} \\[0.125cm]
        \hline\hline
        AdamW & Yes & {2e-8} & 6.50 & 7.62 & 4.63 & 4.60 & 2.76 & 5.96 & 222/0 & 218/0 & 208/1 & 171/5 \\[0.125cm]
        AdamW & Yes & {1e-7} & 9.26 & 11.38 & 6.06 & 6.45 & 3.51 & 9.49 & 222/0 & 214/3 & 188/18 & 132/63 \\[0.125cm]
        \hline
        \parbox[t]{2mm}{\multirow{4}{*}{\rotatebox[origin=c]{90}{Adamax}}}
        & Yes & {2e-8} & 0.81 & 1.14 & 0.62 & 0.63 & 0.75 & -0.22 & 0/0 & 0/0 & 1/0 & 1/0 \\[0.125cm]
        & No & {2e-8} & 6.40 & 7.50 & 4.25 & 4.63 & 3.01 & 6.40 & 224/0 & 220/0 & 214/1 & 178/4 \\[0.125cm]
        & Yes & {1e-7} & 6.67 & 7.65 & 4.67 & 4.68 & 3.51 & 6.40 & 222/0 & 217/0 & 205/2 & 171/7 \\[0.125cm]
        & No & {1e-7} & 6.46 & 7.60 & 4.67 & 5.06 & 3.51 & 6.40 & 222/0 & 217/0 & 198/2 & 157/12 \\
        \bottomrule
    \end{tabular}
    \caption{Percentage improvement over the untuned $E5$ model.  O, M and LR represent the choice of optimizer, whether or not momentum was used and the learning rate, respectively.  All the models here are tuned with a batch size of $14$, margin $0.1$, and a frozen embedding block.  Adamax with no momentum corresponds to choosing $\beta_1 = \beta_2 = 0$ for the optimizer parameters.} 
    \label{tab:optim}
\end{table*}

From the table, the default version of Adamax (Adamax with momentum) has a nearly negligible effect on the model when used with a small learning rate, suggesting that this particular configuration for the optimizer forces the model weights to change very slowly.  When momentum is switched off, the model weights change enough to improve the overall performance in both English and other languages.  Continuing down to the bottom row, if we turn up the learning rate to a higher value, the model weights begin to change more significantly which brings about less improvement in the model's multilingual capacity (still an improvement nonetheless), but maintains the same improvement on English.  Overall, going from the first row to the last row (for Adamax), we transition from a point in model weight space where performance on all languages can be enhanced or preserved to a point which is better suited for the English-only task defined in tuning.


\end{document}